\documentclass[]{interact}

\usepackage{epstopdf}% To incorporate .eps illustrations using PDFLaTeX, etc.
\usepackage{xurl}
\usepackage[caption=false]{subfig}% Support for small, `sub' figures and tables

\usepackage{dsfont}
\usepackage{float}
\usepackage[dvipsnames]{xcolor}

\usepackage[numbers,sort&compress]{natbib}% Citation support using natbib.sty
\bibpunct[, ]{[}{]}{,}{n}{,}{,}% Citation support using natbib.sty
\renewcommand\bibfont{\fontsize{10}{12}\selectfont}% Bibliography support using natbib.sty

\usepackage[ruled,vlined]{algorithm2e}
\SetKwInput{KwInput}{Input}
\SetKwInput{KwOutput}{Output}

\usepackage[bookmarks]{hyperref}
\hypersetup{colorlinks, linkcolor=blue, citecolor=blue,
urlcolor=blue, plainpages=false, pdfwindowui=false,
pdfstartview={FitH}, pdftitle={A}}

\usepackage{comment}
\usepackage{tikz} 
\usepackage{svg}
\theoremstyle{plain}% Theorem-like structures provided by amsthm.sty

\theoremstyle{definition}

\theoremstyle{remark}
\newtheorem{remark}{Remark}

\numberwithin{equation}{section}
\numberwithin{remark}{section}

\newcommand{\indexdata}{b}
\usepackage{makecell}
\newcommand\independent{\protect\mathpalette{\protect\independenT}{\perp}}
\def\independenT#1#2{\mathrel{\rlap{$#1#2$}\mkern2mu{#1#2}}}

\newcommand{\neginc}{\zeta} % opposé de l'incrément

\DeclareMathOperator*{\argmin}{argmin}
\DeclareMathOperator*{\vecteur}{vec}
\DeclareMathOperator*{\MAT}{MAT}

\usepackage{mathtools}

\def\huy#1{{\color{RubineRed}#1}}

\begin{document}

%\articletype{ARTICLE TEMPLATE}% Specify the article type or omit as appropriate

\title{Analysis and Comparison of Two-Level KFAC Methods for Training Deep Neural Networks}

\author{
\name{Abdoulaye \textsc{Koroko}\textsuperscript{a,b}\thanks{Corresponding author: \email{abdoulayekoroko@gmail.fr, abdoulaye.koroko@ifpen.fr}}, Ani \textsc{Anciaux-Sedrakian}\textsuperscript{a}, Ibtihel \textsc{Ben Gharbia}\textsuperscript{a}, Valérie \textsc{Garès}\textsuperscript{c}, Mounir \textsc{Haddou}\textsuperscript{c} and Quang Huy \textsc{Tran}\textsuperscript{a}}
\affil{\textsuperscript{a}IFP Energies nouvelles, 1 et 4 avenue de Bois Préau, 92852 Rueil-Malmaison Cedex, France\\
\textsuperscript{b}CentraleSupélec - Université Paris-Saclay, 3 Rue Joliot Curie, 91190 Gif-sur-Yvette, France\\
\textsuperscript{c}Univ. Rennes, INSA, CNRS, IRMAR - UMR 6625, F-35000 Rennes, France}
}

\maketitle

\begin{abstract}
As a second-order method, the Natural Gradient Descent (NGD) has the ability to accelerate training of neural networks. However, due to the prohibitive computational and memory costs of computing and inverting the Fisher Information Matrix (FIM), efficient approximations are necessary to make NGD scalable to Deep Neural Networks (DNNs). Many such approximations have been attempted. The most sophisticated of these is KFAC, which approximates the FIM as a block-diagonal matrix, where each block corresponds to a layer of the neural network. By doing so, KFAC ignores the interactions between different layers. In this work, we investigate the interest of restoring some low-frequency interactions between the layers by means of two-level methods. Inspired from domain decomposition, several two-level corrections to KFAC using different coarse spaces are proposed and assessed. The obtained results show that incorporating the layer interactions in this fashion does not really improve the performance of KFAC. This suggests that it is safe to discard the off-diagonal blocks of the FIM, since the block-diagonal approach is sufficiently robust, accurate and economical in computation time.
\end{abstract}

\begin{keywords}
Deep Neural Networks; Natural Gradient Descent; Kronecker Factorization; Two-Level Preconditioning
\end{keywords}

\section{Introduction}
 Deep learning has achieved tremendous success in many fields such as computer vision \cite{Krizhevsky2012,HeEtal2016}, speech recognition \cite{sak2014,Sercu2016}, and natural language processing \cite{Bahdanau2014,Gehring2017}, where its models have produced results comparable to human performance. This was made possible thanks not only to parallel computing resources but also to adequate optimization algorithms, the development of which remains a major research area. Currently, the Stochastic Gradient Descent (SGD) method \cite{RobbinsMonro1951} and its variants \cite{Polyak1964SomeMO,Nesterov1983AMF} are the workhorse methods for training DNNs. Their wide adoption by the machine learning community is justified by their simplicity and their relativeley good behavior on many standard optimization problems. Nevertheless, almost all optimization problems arising in deep learning are non-linear and highly non-convex. In addition, the landscape of the objective function may contain
huge variations in curvature along different directions \cite{martensthesis}. This leads to many challenges in DNNs training, which limit the effectiveness of first-order methods like SGD.

\subsection{Approximations of the FIM in NGD methods}
By taking advantage of curvature information, second-order methods can overcome the above-mentioned difficulties and speed up the training of DNNs. In such methods, the gradient is rescaled at each iteration with the inverse of a curvature matrix $C$, whose role is to capture information on the local landscape of the objective function. Several choices of $C$ are available: the well-known Hessian matrix, the Generalized Gauss-Newton matrix (GGN) \cite{Schraudolph2002}, the FIM \cite{Amari1998} or any positive semi-definite approximation of these matrices. The advantage of the GGN and FIM over the Hessian is that they are always positive semi-definite, which is not always guaranteed for the Hessian. 
Despite their theoretical superiority, second-order methods are unfortunately not practical for training DNNs. This is due to the huge computational and memory requirements for assembling and inverting the curvature matrix $C$. Several paradigms have therefore been devised to approximate the curvature matrix of DNNs. For example, the Hessian-free approach (HF) \cite{martens2010} eliminates the need to store $C$ by using a Krylov subspace-based Conjugate Gradient (CG) method to solve the linear system involving $C$. While this approach is memory effective, it remains time-consuming, since one must run at each iteration several steps of CG to converge. Another existing approach is the family of quasi-Newton methods \cite{broyden1970,fletcher1970,goldfard1970,shanno1970,nocedal1989} that rely only on gradient information to build a low-rank approximation to the Hessian. Other popular approximations to the curvature matrix are Adagrad \cite{adagrad}, RMSprop \cite{rmsprop}, and Adam \cite{KingmaBa2015} which develop diagonal approximations to the empirical FIM. Despite their ease of implementation and scalability to DNNs, both low-rank and diagonal approximations throw away a lot of information and, therefore, are in general less effective than a well-tuned SGD with momentum.

More advanced and sophisticated approximations that have sparked great enthusiasm are the family of Kronecker-factored curvature (KFAC) methods. Evolving from earlier works \cite{Heskes2000,LRMB2008,PascanuBengio2013,PZK2014,olivier2015}, KFAC methods \cite{MartensGrosse2015,GrosseMartens2016,MBJ2018} exploit the network structure to obtain a block-diagonal approximation to the FIM. Each block corresponds to a layer and is further approximated by the Kronecker product of two smaller matrices, cheap to store and easy to invert via the formula $(A\otimes B)^{-1}=A^{-1}\otimes B^{-1}$.
Owing to this attractive feature, KFAC has received a lot of attention and many endeavors have been devoted to improving it. In \cite{BaGrosseMartens2017,OTUNYM2019}, distributed versions of KFAC were demonstrated to perform well in large-scale settings. The EKFAC method \cite{GLBBV2018} refines KFAC by rescaling the Kronecker factors with a diagonal variance computed in a Kronecker-factored eigenbasis. The TKFAC method \cite{TKFAC} preserves a trace-invariance relationship between the approximate and the exact FIM. 
By removing KFAC's assumption on the independence between activations and pre-activation derivatives, more rigorous Kronecker factorizations can be worked out \cite{Koroko-et-al}  based on minimization of various errors in the Frobenius norm. 
Beyond the FIM, the idea of Kronecker factorization can also be extended to the Hessian matrix of DNNs as in KBFGS \cite{GRB2020}, where the computational complexity is alleviated by approximating the inverse of the Kronecker factors with low-rank updates, as well as the GGN matrix of Multi-Layer Perceptrons (MLP), as shown in \cite{GRB2020}. KFAC has also been deployed successfully in the context of Bayesian deep learning \cite{kfacbay}, deep reinforcement learning \cite{kfacrl} and Laplace approximation \cite{ritter2018a}.

\subsection{Enhancement of KFAC by rough layer interaction}
For computation and memory purposes, KFAC as well as all related variants use only a block-diagonal approximation of the curvature matrix, where each block corresponds to a layer. This results in a loss of information about the correlations between different layers. The question then naturally arises as to whether it is worth trying to recover some of the lost information in hope of making the approximate FIM closer to the true one, thus improving the convergence speed of the optimizer without paying an excessive price.

To this question, Tselepidis et al. \cite{twolevels} provided an element of answer by considering a ``coarse'' correction to the inverse of the approximate FIM. This additional term is meant to represent the interaction between layers at a ``macroscopic'' scale, in contrast with the ``microscopic'' scale of the interaction between neurons inside each layer. Their approach proceeds by formal analogy with the two-level preconditioning technique in domain decomposition \cite{Jolivet2015}, substituting the notion of layer for that of subdomain. The difference with domain decomposition, however, lies in the fact that the matrix at hand does not stem from the discretization of any PDE system, and this prevents the construction of coarse spaces from being correctly guided by any physical sense. Notwithstanding this concern, some ready-made recipes can be blindly borrowed from two-level domain decomposition. In this way, Tselepidis et al. \cite{twolevels} reached a positive conclusion regarding the advisability of enriching the approximate FIM with some reduced information about interactions between layers. Nevertheless, their coarse correction is objectionable in some respects, most notably because of inconsistency in the formula for the new matrix (see \S\ref{sec:Two-level} for a full discussion), while for the single favorable case on which is based their conclusion, the network architecture selected is a little too simplistic (see \S\ref{sec:conclusion} for details). Therefore, their claim should not be taken at face value.

Although he did not initially intend to look at the question as formulated above, Benzing \cite{Benzing2022GradientDO} recently brought another element of answer that runs counter to the former. By carefully comparing KFAC and the exact natural gradient (as well as FOOF, a method of his own), he came to the astonishingly counterintuitive conclusion that KFAC outperforms the exact natural gradient in terms of optimization performance. In other words, there is no benefit whatsoever in trying to embed any kind of information about the interaction between layers into the curvature matrix, since even the full FIM seems to worsen the situation. While one may not be convinced by his heuristical explanation (whereby KFAC is argued to be a first-order method), his numerical results eloquently speak for themselves. Because Benzing explored a wide variety of networks, it is more difficult to mitigate his findings.

In light of these two contradictory sets of results, we undertook this work in an effort to clarify the matter. To this end, our objective is first to design a family of coarse corrections to KFAC that do not suffer from the mathematical flaws of Tselepidis et al.'s one. This gives rise to a theoretically sound family of approximate FIMs that will next be compared to the original KFAC. This leads to the following outline for the paper. In \S\ref{sec:preliminaries}, we introduce notations and recall essential prerequisites on the network model, the natural gradient descent, and the KFAC approximation. In \S\ref{sec:Two-level}, after pointing out the shortcomings of Tselepidis et al.'s corrector, we put forward a series of two-level KFAC methods, the novelty of which is their consistency and their choices of the coarse space. In \S\ref{sec:experiments}, we present and comment on several experimental results, which include much more test cases and analysis in order to assess the new correctors as fairly as possible. Finally, in \S\ref{sec:conclusion}, we summarize and discuss the results before sketching out some prospects.

\section{Backgrounds on the second-order optimization framework}\label{sec:preliminaries}

\subsection{Predictive model and its derivatives}
We consider a feedforward neural network $f_{\theta}$, containing $\ell$ layers and parametrized by 
\begin{equation}
\theta = [\text{vec}(W_1)^T,\text{vec}(W_2)^T, \hdots, \text{vec}(W_{\ell})^T]^T \in\mathbb{R}^p,
\end{equation}
where $W_i$ is the weights matrix associated to layer $i$ and ``vec'' is the operator that vectorizes a matrix by stacking its columns together. We also consider a training data 
\[
\mathcal{U} = \big\{ (x^{(1)},y^{(1)}), \, (x^{(2)},y^{(2)}), \,\hdots , \, (x^{(n)},y^{(n)})\: | \:  (x^{(\indexdata)},y^{(\indexdata)}) \in \mathbb{R}^{d_x\times d_y} , \: 1\leq b\leq n \big\}
\]
and a loss function $L(y,z)$ which measures the discrepancy between the actual target $y$ and the network's prediction $z=f_{\theta}(x)$ for a given input-target pair $(x,y) \in \mathcal{U}$. The goal of the training problem
\begin{equation}\label{eq:trainingpb}
\argmin_{\theta \in \mathbb{R}^p} h(\theta) := \frac{1}{n}\sum_{\indexdata=1}^n L(y^{(\indexdata)},f_{\theta}(x^{(\indexdata)}))
\end{equation}
is to find the optimal value of $\theta$ that minimizes the empirical risk $h(\theta)$.
In the following, we will designate by $\mathcal{D}v=\nabla_{v}L$ the gradient of the loss function with respect to any variable $v$. 
Depending on the type of network, its output and the gradient of the loss are computed in different ways. Let us describe the calculations for two types of networks.

\paragraph*{MLP (Multi-Layer Perceptron).} Given an input $x\in\mathbb{R}^{d_x}$, the network computes its output $z=f_{\theta}(x) \in \mathbb{R}^{d_y}$ through the following sequence, known as forward-propagation: starting from $a_0:= x$, we carry out the iterations
\begin{equation}
s_i = W_i \Bar{a}_{i-1}, \qquad a_i=\sigma_i(s_i),  \qquad \text{for}\; i \; \text{from} \; 1\; \text{to} \; \ell,
\end{equation}
where $\Bar{a}_{i-1}\in\mathbb{R}^{d_{i-1}+1}$ is $a_{i-1}$ concatenated with $1$ in order to capture the bias and $\sigma_i$ is the activation function at layer $i$. Here, $W_i \in\mathbb{R}^{d_i\times (d_{i-1}+1)}$, with $d_i$ the number of neurons in layer $i$. The sequence is terminated by $z := a_\ell$. Note that the total number of parameters is necessarily $p= \sum_{i=1}^{\ell} d_i (d_{i-1} + 1)$.

The gradient of the loss with respect to the parameters is computed via the back-propagation algorithm: starting from
$\mathcal{D}a_{\ell} = \partial_z L(y,z=a_\ell)$, we perform
\begin{equation}
g_i= \mathcal{D}a_i \odot \sigma'_i(s_i), \quad
 \mathcal{D}W_{i}= g_{i}\Bar{a}_{i-1}^T, \quad
 \mathcal{D}a_{i-1} = W_i^T g_i,\quad \text{for}\; i \; \text{from} \; \ell\; \text{to} \; 1,
\end{equation}
where the special symbol $g_i := \mathcal{D}s_i$ stands for the preactivation derivative. Note that, in the formula for $\mathcal{D}a_{i-1}$, the last row of $W_i^T$ should be removed so that the product in the right-hand side belongs to $\mathbb{R}^{d_{i-1}}$.

\paragraph*{CNN (Convolutional Neural Network).} The calculation is governed by the same principle as for MLP, but the practical organization slightly differs from that of MLP. In a convolution layer, the input, which is an image with multiple channels is convolved with a set of filters to produce an output image containing multiple channels. In order to speed up computations, traditional convolution operations are reshaped into matrix-matrix or matrix-vector multiplications using the unrolling approach \cite{Chellapilla2006HighPC}, whereby the input/output data are copied and rearranged into new matrices (see Figure \ref{fig:CNN}).
\begin{figure}[h]
    \centering
    \includegraphics[scale=0.35]{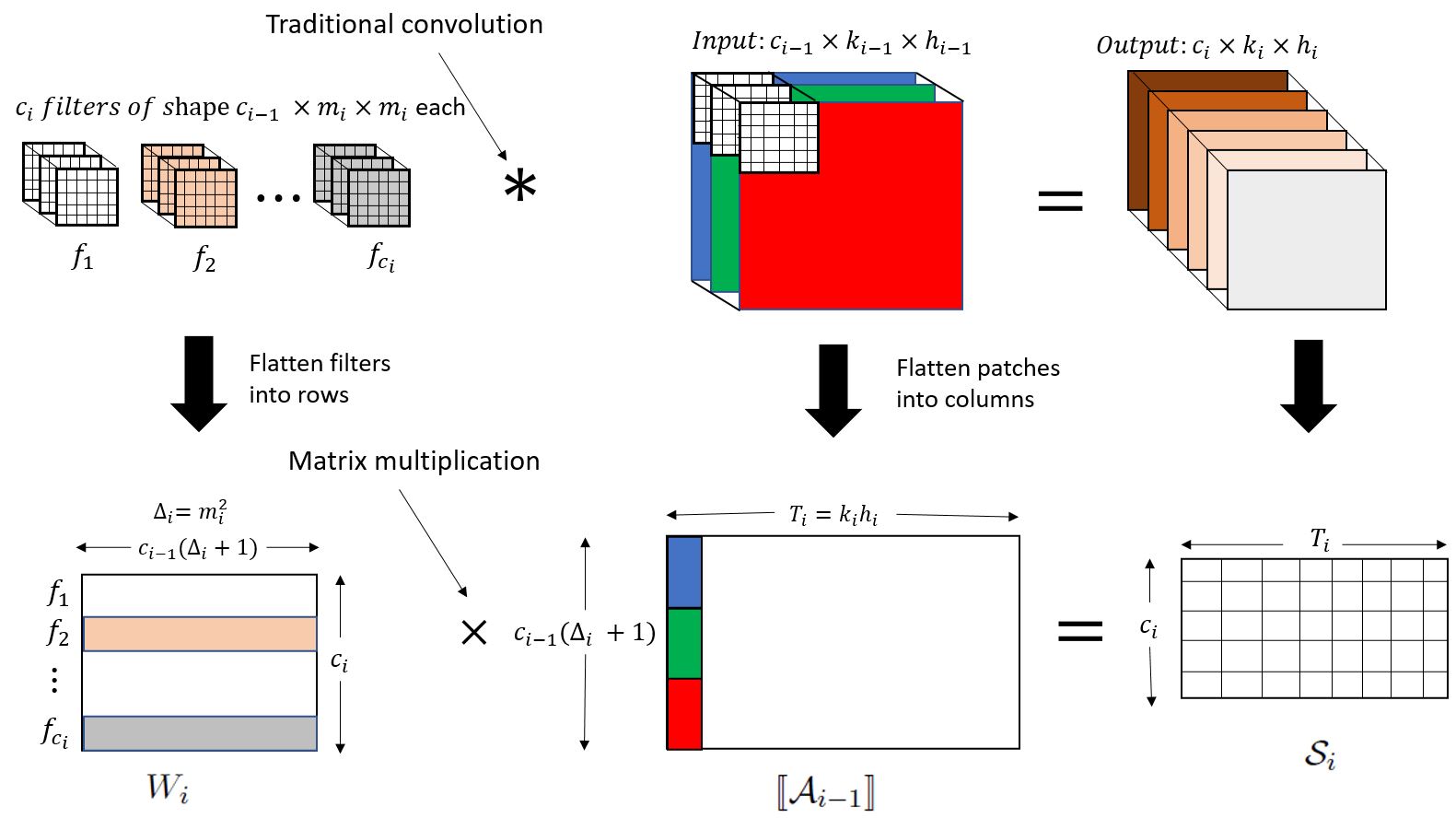}
    \caption{Traditional convolution turned into matrix-matrix multiplication with the \emph{unrolling} approach.}
    \label{fig:CNN}
\end{figure}

Assume that layer $i$ which receives input $\mathcal{A}_{i-1}\in \mathbb{R}^{c_{i-1}\times T_{i-1}}$, where $T_{i-1}$ denotes the number of spatial locations and $c_{i-1}$ the number of channels. Considering $c_i$ filters, each of which involves $\Delta_i$ coefficients, we form a weight matrix $W_i$ of shape $c_i\times (c_{i-1}\Delta_i+1)$, where each row corresponds to a single filter flattened into a vector. Note that the additional $1$ in the column dimension of $W_i$ is required for the bias parameter. Around each position $t\in \{1,\ldots, T_{i-1}\}$,  we define the local column vector $a_{i-1,t}\in \mathbb{R}^{c_{i-1}\Delta_i}$ by extracting the patch data from $\mathcal{A}_{i-1}$ (cf. \cite{GrosseMartens2016} for explicit formulas). The output $\mathcal{A}_i \in \mathbb{R}^{c_i\times T_i}$ is computed by the forward-propagation:
%$\forall{t} \in \{1,\hdots,T_i\}$, let $\Bar{a}_{i-1,t} \in \mathbb{R}^{c_{i-1}\Delta_i+1}$ be the vector formed with the patch (the same shape as a filter) extracted from $\mathcal{A}_{i-1}$ around $t$. Then, 
for $t\in\{1,\ldots,T_i\}$, the $t$-th column  $\widetilde{a}_{i,t}$ of $\mathcal{A}_i$ is given by
\begin{equation}
s_{i,t}=W_i\Bar{a}_{i-1,t}, \qquad \widetilde{a}_{i,t} = \sigma_i (s_{i,t}),
\end{equation}
where $\Bar{a}_{i-1,t}\in\mathbb{R}^{c_{i-1}\Delta_i+1}$ is $a_{i-1}$ concatenated with $1$ in order to capture the bias.
In matrix form, let $[\![\mathcal{A}_{i-1}]\!] \in \mathbb{R}^{(c_{i-1}\Delta_i+1)\times T_i}$ be the matrix whose $t$-th column is $\Bar{a}_{i-1,t}$. Then,
\begin{equation}
\mathcal{S}_i = W_i [\![\mathcal{A}_{i-1}]\!], \qquad 
\mathcal{A}_{i}=\sigma_i(\mathcal{S}_i).
\end{equation}

The gradient of the loss with respect to the parameters of layer $i$ is computed as via the back-propagation formulas
\begin{equation}
g_{i,t}= \mathcal{D}\widetilde{a}_{i,t} \odot \sigma'_i(s_{i,t}),
\quad \mathcal{D}W_{i} 
%=  \sum_{t=1}^{T_i}\frac{\partial L}{\partial s_{i,t}}\frac{\partial s_{i,t}}{W_i}
           =   \sum_{t=1}^{T_i} g_{i,t} \Bar{a}_{i-1,t}^T,
\quad 
\mathcal{D}a_{i-1,t} = W_i^T g_{i,t},
\end{equation}
for $t\in T_i$, where the special symbol $g_{i,t} := \mathcal{D}s_{i,t}$ stands for the preactivation derivative.

In all cases (MLP or CNN), the gradient $\nabla_\theta L = \mathcal{D}\theta$ of the loss with respect to whole parameter $\theta$ is retrieved as
\begin{equation}
\mathcal{D}\theta = [\text{vec}(\mathcal{D}W_1)^T,\text{vec}(\mathcal{D}W_2)^T, \ldots, \text{vec}(\mathcal{D}W_{\ell})^T]^T.
\end{equation}
A general descent method to solve the training problem \eqref{eq:trainingpb} is based on the iterates
\begin{equation}\label{eq:algogeneral}
    \theta_{k+1} = \theta_k - \alpha_k [C(\theta_k)]^{-1}\nabla_{\theta} h({\mathcal{S}_k,\theta_{k}}), 
\end{equation}
where $\alpha_k>0$ is the learning rate,
\begin{equation}
\nabla_{\theta} h({\mathcal{S}_k,\theta_{k}}) = \frac{1}{|\mathcal{S}_k|}
\sum_{ (x^{(\indexdata)},y^{(\indexdata)}) \in \mathcal{S}_k} \nabla_\theta L(y^{(\indexdata)},f_{\theta_k}(x^{(\indexdata)}))
\end{equation}
is a batch approximation of the full gradient $\nabla_\theta h(\theta_k) = \frac{1}{n}\sum_{\indexdata=1}^n \nabla_\theta L(y^{(\indexdata)}, f_{\theta_k}(x^{(\indexdata)}))$
on a random subset $\mathcal{S}_{k} \subset \mathcal{U}$, and $C(\theta_k)$ is an invertible matrix which depends on the method being implemented.

\subsection{Natural Gradient Descent}
The \emph{Natural Gradient Descent} (NGD) is associated with a particular choice for matrix $C(\theta_k)$, which is well-defined under a mild assumption.
\paragraph*{Hypothesis on the loss function.} From now on, we take it for granted that there exists a probability density $\wp(y|z)$ on $y\in \mathbb{R}^{d_y}$ such that, up to an additive constant $\nu$, the loss function $L(y,z)$ takes the form 
\begin{equation}
    L(y,z) =-\log \wp(y|z) + \nu.
\end{equation}
For instance, if the elementary loss corresponds to the least-squares function
\begin{subequations}
\begin{equation}
    L(y,z) = \frac{1}{2} \| y - z \|_2^2,
\end{equation}
then we can take the normal density
\begin{equation}
    \wp(y|z) = (2\pi)^{-d_y/2} \exp(- \textstyle\frac{1}{2} \|y- z\|_2^2), 
\end{equation}
so that
\begin{equation}
L(y,z) = - \log \wp(y|z) - \frac{d_y}{2} \log(2\pi).
\end{equation}
\end{subequations}

Introduce the notation $p(y|x,\theta) = \wp(y| f_\theta(x))$. Then, the composite loss function
\begin{equation}
    L(y, f_\theta(x)) = -\log p(y|x,\theta)
\end{equation}
derives from the density function $p(y|x,\theta)$ of the model's conditional predictive distribution $P_{y|x}(\theta)$. As shown above, $P_{y|x}(\theta)$ is multivariate normal for the standard square loss function. It can also be proved that $P_{y|x}(\theta)$ is multinomial for the cross-entropy one. The learned distribution is therefore $\mathsf{P}_{x,y}(\theta)$ with density 

\begin{equation}
    \mathsf{p}(x,y|\theta) = q(x)p(y|x,\theta) ,
\end{equation}
where $q(x)$ is the density of data distribution $Q_x$ over inputs $x\in\mathbb{R}^{d_x}$.
%\vspace{0.8cm}

\paragraph*{Fisher Information Matrix.} The NGD method \cite{Amari1998} is defined as the generic algorithm \eqref{eq:algogeneral} in which $C(\theta)$ is set to the {\em Fisher Information Matrix} (FIM)
\begin{subequations}\label{eq:FIM}
\begin{align}
 F(\theta) &= \mathbb{E}_{(x,y)\sim \mathsf{P}_{x,y}(\theta)} \{ \nabla_{\theta} \log \mathsf{p}(x,y|\theta) [\nabla_{\theta} \log \mathsf{p}(x,y|\theta)]^T \} \\
          &= \mathbb{E}_{(x,y)\sim \mathsf{P}_{x,y}(\theta)} \{ \mathcal{D}\theta (\mathcal{D}\theta)^T\}\\
          &= \text{cov}(\mathcal{D}\theta,\mathcal{D}\theta),
\end{align}
\end{subequations}
where $\mathbb{E}_{(x,y)\sim \mathsf{P}_{x,y}(\theta)}$ denotes the expectation taken over the prescribed distribution $\mathsf{P}_{x,y}(\theta)$ at a fixed $\theta$. To alleviate notations, we shall write $\mathbb{E}$ instead of $\mathbb{E}_{(x,y)\sim \mathsf{P}_{x,y}(\theta)}$ from now on. Likewise, we shall write $F$ instead of $F(\theta)$ or $F(\theta_k)$.

By definition \eqref{eq:FIM}, the FIM is a covariance matrix and is therefore always positive semi-definite. However, for the iteration
\begin{equation}\label{eq:algoNGD}
     \theta_{k+1} = \theta_k - \alpha_k F^{-1}\nabla_{\theta} h(\mathcal{S}_k,\theta_k)
\end{equation}
to be well-defined, $F$ has to be invertible. This is why, in practice, $C(\theta_k)$ will be taken to be a regularized version of $F$ under the form
\begin{equation}\label{eq:FIMreg}
F_{\bullet} = F + \lambda I_p, \qquad \lambda >0.
\end{equation}
The actual NGD iteration is therefore
\begin{equation}\label{eq:algoNGDreg}
     \theta_{k+1} = \theta_k - \alpha_k F_{\bullet}^{-1}\nabla_{\theta} h(\mathcal{S}_k,\theta_k) .
\end{equation}

In the space of probability distributions $\mathsf{P}_{x,y}(\theta)$ equipped with the {\em Kullback-Leibler} (KL) divergence, the FIM represents the local quadratic approximation of this induced metric, in the sense that for a small vector $\delta \in \mathbb{R}^p$, we have
\begin{equation}
    \mathrm{KL} [\mathsf{P}_{x,y}(\theta) \, \| \, \mathsf{P}_{x,y}(\theta+\delta) ] = \frac{1}{2}\delta^T F \delta+O(\|\delta\|^3),
\end{equation}
where $\mathrm{KL}[\mathsf{P}||\mathsf{Q}]$ is the KL divergence between the distributions $\mathsf{P}$ and $\mathsf{Q}$
As a consequence, the unregularized NGD can be thought of as the steepest descent in this space of probability distributions \cite{AmariNagaoka2000}.

\paragraph*{Advantages and drawbacks.} By virtue of this geometric interpretation, the NGD has the crucial advantage of being intrinsic, that is, invariant with respect to invertible reparameterizations. Put another way, the algorithm will produce matching iterates regardless of how the unknowns are transformed. This is useful in high-dimensional cases where the choice of parameters is more or less arbitrary.

Strictly speaking, invariance with respect to parameters only occurs at the continuous level, i.e., in the limit of $\alpha_k\rightarrow 0$. This minor drawback does not undermine the theoretical soundness of the method. In fact, the real drawback of the NGD \eqref{eq:FIM}--\eqref{eq:algoNGD} lies in the cost of computing and inverting the Fisher matrix. This is why it is capital to consider suitable approximations of the FIM such as KFAC (cf. \S\ref{sse:KFAC}).

Finally and anecdotically, the NGD method can also be viewed as an approximate Newton method since the FIM and the GGN matrix are equivalent when the model predictive distribution  $P_{y|x}(\theta)$ belongs to exponential family distributions \cite{Martens2014}.

\subsection{KFAC approximation of the FIM}\label{sse:KFAC}
From equation \eqref{eq:FIM}, the FIM can be written as a block-diagonal matrix
\begin{subequations}
\begin{equation}
F =  \mathbb{E} [\mathcal{D}\theta (\mathcal{D}\theta)^T ]
 =   \begin{bmatrix} 
    F_{1,1}  & \dots & F_{1,\ell} \\
    \vdots  & & \vdots \\
     F_{\ell,1}  & \dots & F_{\ell,\ell}
    \end{bmatrix} \in \mathbb{R}^{p\times p},
\end{equation}
in which each block $F_{i,j}$ is given by
\begin{equation}
F_{i,j} =   \mathbb{E}[\text{vec}(\mathcal{D}W_i)\text{vec}(\mathcal{D}W_j)^T].
\end{equation}
\end{subequations}
One can interpret $F_{i,i}$  as being second-order statistics of weight derivatives of layer $i$, and $F_{i,j}$, $i \neq j$ as representing the interactions between layer $i$ and $j$.

The KFAC method \cite{MartensGrosse2015} is grounded on the following two assumptions to provide an efficient approximation to the FIM that is convenient for training DNNs. 
\begin{enumerate}
\item The first one is that there are no interactions between two different layers, i.e., $F_{i,j}=0$ for $i\neq j$. This results in the block-diagonal approximation
\begin{equation}
F\approx \widetilde{F} = \text{diag}(F_{1,1},F_{2,2},\hdots, F_{\ell,\ell}) 
\end{equation}
for the FIM.
At this step, computing the inverse of $\widetilde{F}$ is equivalent to computing the inverses of diagonal blocks $F_{i,i}$. Nevertheless, because the diagonal blocks $F_{i,i}$ can be very large (especially for DNNs with large layers), this first approximation remains insufficient. 
\item The second one comes in support of the first one and consists in factorizing each diagonal block $F_{i,i}$ as a Kronecker product of two smaller matrices, namely,
\begin{equation}
F_{i,i}\approx [F_{\text{KFAC}}]_{i,i} = A_i\otimes G_i.
\end{equation}
where the Kronecker product between $A \in \mathbb{R}^{m_A\times n_A}$ and $B\in \mathbb{R}^{m_B\times n_B}$ is the matrix of size $m_A m_B\times n_A n_B$ given by
\begin{equation}
A\otimes B =  \begin{bmatrix} 
    A_{1,1}B  & \dots & A_{1,n_A}B \\
    \vdots &  & \vdots \\
   \, A_{m_A,1}B & \dots   & A_{m_A,n_A}B \;
    \end{bmatrix} .
\end{equation}
\end{enumerate}
Now, depending on the type of the layer, the computation of the Kronecker factors $A_i$ and $G_i$ may require different other assumptions.

\paragraph*{MLP layer.} When layer $i$ is an MLP, the block $F_{i,i}$ is given by 
\begin{align}
  F_{i,i} & = \mathbb{E}[\text{vec}(\mathcal{D}W_i)\text{vec}(\mathcal{D}W_j)^T] \nonumber \\
          & = \mathbb{E}[\text{vec}(g_{i}\Bar{a}_{i-1}^T)\text{vec}(g_{i}\Bar{a}_{i-1}^T)^T] \nonumber\\
          & = \mathbb{E}[(\Bar{a}_{i-1}\otimes g_i)(\Bar{a}_{i-1}\otimes g_i)^T] \nonumber\\
          & = \mathbb{E}[\Bar{a}_{i-1}\Bar{a}_{i-1}^T\otimes g_ig_i^T] .
\end{align}
From the last equality, if one assumes that activations and pre-activation derivatives are independent, that is, $ a_{i-1} \independent g_i$, then $F_{i,i}$ can be factorized as
\begin{equation}
F_{i,i} \approx [F_{\text{KFAC}}]_{i,i}=\mathbb{E}[\Bar{a}_{i-1}\Bar{a}_{i-1}^T]\otimes\mathbb{E}[ g_i g_i^T] \, =: A_i\otimes G_i,
\end{equation}
with $A_i = \mathbb{E}[\Bar{a}_{i-1}\Bar{a}_{i-1}^T]$ and $G_i=\mathbb{E}[ g_i g_i^T]$.

\paragraph*{Convolution layer.} With such a layer, $F_{i,i}$ is written as 
\begin{align}
F_{i,i} &= \mathbb{E} \big[\text{vec}(\mathcal{D}W_i)\text{vec}(\mathcal{D}W_j)^T \big] \nonumber \\
        &= \mathbb{E} \bigg[ \text{vec} \bigg(\sum_{t=1}^{T_i} g_{i,t} \Bar{a}_{i-1,t}^T \bigg) \text{vec} \bigg( \sum_{t=1}^{T_i} g_{i,t} \Bar{a}_{i-1,t}^T \bigg)^T \bigg] \nonumber\\
        &= \mathbb{E} \bigg[ \sum_{t=1}^{T_i} \sum_{t^{\prime}=1}^{T_i}(\Bar{a}_{i-1,t}\otimes g_{i,t}) (\Bar{a}_{i-1,t^{\prime}}\otimes g_{i,t^{\prime}})^T \bigg] \nonumber\\
         &= \mathbb{E} \bigg[ \sum_{t=1}^{T_i} \sum_{t^{\prime}=1}^{T_i}\Bar{a}_{i-1,t}\Bar{a}_{i-1,t^{\prime}}^T\otimes g_{i,t} g_{i,t^{\prime}}^T \bigg] \nonumber\\
        &= \mathbb{E} \bigg[ \sum_{t=1}^{T_i} \sum_{t^{\prime}=1}^{T_i}\Omega_i(t,t^{\prime})\otimes \Gamma_i(t,t^{\prime}) \bigg],
\end{align}
with $\Omega_i(t,t^{\prime})=\Bar{a}_{i-1,t}\Bar{a}_{i-1,t^{\prime}}^T$ and $\Gamma_i(t,t^{\prime})=g_{i,t} g_{i,t^{\prime}}^T$.
In order to factorize $F_{i,i}$ into Kronecker product of two matrices, Grosse and Martens \cite{GrosseMartens2016} resort to three hypotheses. First, similarly to MLP layers, activations and pre-activation derivatives are assumed to be independent. Secondly, postulating spatial homogeneity, the second-order statistics of the activations and pre-activation derivatives at any two spatial locations $t$ and $t^{\prime}$ depend only on the difference $t-t^{\prime}$. Finally, the pre-activation derivatives at any two distinct spatial locations are declared to be uncorrelated, i.e., $\Gamma_i(t,t^{\prime})=0 \: \text{for}\: t\neq t^{\prime}$. Combining these three assumptions yields the approximation
\begin{equation}
F_{i,i} \approx [F_{\text{KFAC}}]_{i,i} = \mathbb{E} \bigg[ \sum_{t=1}^{T_i}\Omega_i(t,t) \bigg] \otimes
\frac{1}{T_i}\mathbb{E} \bigg[\sum_{t=1}^{T_i}\Gamma_i(t,t) \bigg] \, =: A_i\otimes G_i,
\end{equation}
with $A_i=\mathbb{E} \big[ \sum_{t=1}^{T_i}\Omega_i(t,t) \big]$ and $G_i= \frac{1}{T_i}\mathbb{E} \big[\sum_{t=1}^{T_i}\Gamma_i(t,t) \big]$.
%\vspace{0.5cm}

\begin{remark}
It should be mentioned that, in the same spirit, a KFAC-type approximation has been developed for the RNN (Recurrent Neural Network), but with much more assumptions. In this work, we do not consider recurrent layers. The readers interested in KFAC for RNN are referred to \cite{MBJ2018}.
\end{remark}

Going back to MLP and CNN layers, the matrices $A_i$ and $G_i$ are estimated using Monte Carlo method, with a mini-batch $\mathcal{B}=\{(x_1,y_1),\hdots,(x_B,y_B)\}$, where the targets $y_i$'s are sampled from the model predictive distribution $P_{y|x}(\theta)$.
Combining the block-diagonal approximation and Kronecker factorization of each block, the approximate FIM becomes
\begin{equation}\label{eq:KFACdiag}
F\approx F_{\text{KFAC}} = \text{diag}(A_1\otimes G_1, \, A_2\otimes G_2, \, \hdots, \, A_{\ell}\otimes G_{\ell}).
\end{equation}
The descent iteration \eqref{eq:algogeneral} with $C(\theta_k) = F_{\text{KFAC}}(\theta_k)$ is now well suited to training DNNs. Indeed, thanks to the Kronecker product properties $(A\otimes B)^{-1} = A^{-1} \otimes B^{-1}$ and $(A\otimes B)\text{vec}(X)= \text{vec}(BXA^T)$, it is plain that the product
\begin{equation}\label{eq:KFAC}
    F_{\text{KFAC}}^{-1}\nabla_{\theta} h = \begin{bmatrix}
    \text{vec} (G_{1}^{-1}(\nabla_{W_1}h){A}_{1}^{-1}) \\ \vdots \\ 
    \text{vec} (G_{\ell}^{-1}(\nabla_{W_{\ell}}h){A}_{\ell}^{-1})
    \end{bmatrix}
\end{equation}
only requires to store and to invert matrices of moderately small sizes. 

In practice, invertibility  of $F_{\text{KFAC}}$ must be enforced by a regularization procedure. The usual Tikhonov one, by which we consider $F_{\text{KFAC}}+\lambda I, \lambda>0$, instead of $F_{\text{KFAC}}$, is equivalent to adding a multiple of the identity matrix of appropriate size to each diagonal block, i.e., $A_i\otimes G_i + \lambda I_i$. Unfortunately, this breaks down the Kronecker factorization structure of the blocks. To preserve the factorized structure, the authors of KFAC \cite{MartensGrosse2015} advocate a heuristic damping technique in which each Kronecker factor is regularized as
\begin{subequations}
\begin{equation}
[F_{\bullet\,\text{KFAC}}]_{i,i}   = (A_i+\pi_i \lambda^{1/2} I_{A_{\huy{i}}})\otimes (G_i+\pi_i^{-1} \lambda^{1/2} I_{G_{\huy{i}}}),
\end{equation}
where $I_{A_{i}}$ and $I_{G_{i}}$ denote identity matrices of same size as $A_i$ and $G_i$ respectively, and 
\begin{equation}
\pi_i = \sqrt{\frac{\text{tr}(A_{i})/(d_{i-1}+1)}{\text{tr}(G_i)/d_i}} .
\end{equation}
\end{subequations}
The actual KFAC iteration is therefore
\begin{subequations}
\begin{equation}\label{eq:algoKFACreg}
     \theta_{k+1} = \theta_k - \alpha_k F_{\bullet\,\text{KFAC}}^{-1}\nabla_{\theta} h(\mathcal{S}_k,\theta_k) ,
\end{equation}
with
\begin{equation}
    F_{\bullet\,\text{KFAC}} = \text{diag}([F_{\bullet\,\text{KFAC}}]_{1,1} , \, [F_{\bullet\,\text{KFAC}}]_{2,2}, \, \hdots, \, [F_{\bullet\,\text{KFAC}}]_{\ell,\ell}).
\end{equation}
\end{subequations}

\section{Two-level KFAC methods}\label{sec:Two-level}
Henceforth, the learning rate is assumed to be $\alpha_k = 1$. Let
\begin{equation}
\neginc_k = \theta_k - \theta_{k+1} = [C(\theta_k)]^{-1} \nabla_\theta h(\mathcal{S}_k,\theta_k)
\end{equation}
be the negative increment of $\theta$ at iteration $k$ of the generic descent algorithm \eqref{eq:algogeneral}. To further alleviate notations, we shall drop the subscript $k$ and omit the dependence on $\theta_k$. For the regularized NGD \eqref{eq:algoNGDreg}, we have
\begin{equation}\label{eq:sys}
\neginc = F_{\bullet}^{-1} \,\nabla_\theta h,
\end{equation}
while for the regularized KFAC method, we have
\begin{equation}\label{eq:sysKFAC}
 \neginc_{\text{KFAC}} = F_{\bullet\,\text{KFAC}}^{-1} \nabla_\theta h,
\end{equation}
being understood that the matrices are regularized whenever necessary.

We want to build a new matrix $F_{\bullet\,\text{KFAC-2L}}^{-1}$, an augmented version of $F_{\bullet\,\text{KFAC}}^{-1}$, such that the solution
\begin{equation}\label{eq:sysKFAC2L}
\neginc_{\text{KFAC-2L}} = F_{\bullet\,\text{KFAC-2L}}^{-1} \nabla_\theta h,
\end{equation}
is a better approximation to $\neginc$ than $\neginc_{\text{KFAC}}$, namely,
\begin{equation}
\| \neginc_{\text{KFAC-2L}} - \neginc \|_{F} \ll \| \neginc_{\text{KFAC}} - \neginc \|_{F} .
\end{equation}
By ``augmented'' we mean that, at least partially and at some rough scale, $F_{\text{KFAC-2L}}^{-1}$ takes into account the information about layer interactions that was discarded by the block-diagonal approximation KFAC. The basic tenet underlying this initiative is the belief that a more accurate approximation to the NGD solution $\neginc$ at each descent iteration will help the global optimization process to converge faster.

\subsection{Analogy and dissimilarity with domain decomposition}
The construction philosophy of $F_{\bullet\,\text{KFAC-2L}}^{-1}$ proceeds by analogy with insights from domain decomposition. To properly explain the analogy, we first need to cast the matrix $F_{\bullet\,\text{KFAC}}^{-1}$ under a slightly different form.

For each $i\in\{1,\ldots,\ell\}$, let $R_i \in \mathbb{R}^{p_i\times p}$  be the matrix of the restriction operator from $\mathbb{R}^p$, the total space of all parameters, to the subspace of parameters pertaining to layer $i$, whose dimension is $p_i$. In other words, for $(\xi,\eta)\in \{1,\ldots,p_i\} \times \{ 1,\ldots, p\}$,
\begin{equation}\label{eq:restriction}
(R_i)_{\xi\eta} = \begin{cases}
\, 1 & \;\text{if }\; \eta = p_1 + \ldots + p_{i-1} + \xi,\\
\, 0 & \;\text{otherwise}.
\end{cases}
\end{equation}
The transpose $R_i^T \in \mathbb{R}^{p\times p_i}$ then represents the prolongation operator from the subspace of parameters in layer $i$ to the total space of all parameters. Obviously, the $i$-th diagonal block of the regularized FIM can be expressed as
\[
[F_{\bullet}]_{i,i} = R_i F_{\bullet} R_i^T .
\]
If there were no approximation of each diagonal block by a Kronecker product, then the block-diagonal approximation of $F$ would give rise to the inverse matrix
\begin{equation}\label{eq:FIMinvDD}
F_{\bullet\,\text{block-diag}}^{-1} = \sum_{i=1}^{\ell} R_i^T [F_{\bullet}]_{i,i}^{-1} R_i = \sum_{i=1}^{\ell} R_i^T (R_i F_{\bullet} R_i^T)^{-1} R_i  .
\end{equation}
In the case of KFAC, it follows from \eqref{eq:KFACdiag}--\eqref{eq:KFAC} that
\begin{align}\label{eq:KFACinvDD}
F_{\bullet\,\text{KFAC}}^{-1} &= \sum_{i=1}^{\ell} R_i^T [F_{\bullet\,\text{KFAC}}]_{i,i}^{-1} R_i\\[-1mm]
& = \sum_{i=1}^{\ell} R_i^T (A_i+\pi_i \lambda^{1/2} I_{A_{i}})^{-1}\otimes (G_i+\pi_i^{-1} \lambda^{1/2} I_{G_{i}})^{-1} R_i  . \nonumber
\end{align}

In the context of the domain decomposition methods to solve linear systems arising from the discretization of PDEs, the spatial domain of the initial problem is divided into several subdomains. The system is then projected onto the subdomains and the local subproblems are solved independently of each other as smaller systems. In this stage, parallelism can be fully taken advantage of by assigning a processor to each subdomain. This produces a local solution on each subdomain. These local solutions are next combined to create an approximate global solution on the overall domain. Algebraically, the whole process is tantamount to using an inverse matrix of a form similar to \eqref{eq:FIMinvDD}--\eqref{eq:KFACinvDD} either within a Schwarz-like iterative procedure or as a preconditioner \cite{Jolivet2015}. The counterparts of $[F_{\bullet}]_{i,i}^{-1}$ or $[F_{\bullet\,\text{KFAC}}]_{i,i}^{-1}$ are referred to as \emph{local solvers}.

\begin{remark}
The above analogy is not a perfect one. In domain decomposition, the subdomains are allowed (and even recommended!) to overlap each other, so that an unknown can belong to two or more subdomains. In this case, the restriction operators $R_i$ can be much more intricate than the one trivially defined in \eqref{eq:restriction}.

\end{remark}

A well-known issue with domain decomposition methods of the form \eqref{eq:FIMinvDD}--\eqref{eq:KFACinvDD} is the disappointingly slow rate of convergence, which results in a lack of \emph{scalability} \cite{Jolivet2015}: the speed-up factor does not grow proportionally with the number of subdomains (and therefore of processors). The reason is that, as the number of subdomains increases, it takes more iterations for an information local to one subdomain to be propagated and taken into account by the others. The common remedy to this problem is to append a ``coarse'' correction that enables subdomains to communicate with each other in a faster way. The information exchanged in this way is certainly not complete, but only concerns the low frequencies.

\begin{remark}
In domain decomposition, there is a physical problem (represented by the PDE at the continuous level) that serves as a support for the mathematical and numerical reasoning. This is not the case here, where we have to think in a purely algebraic way.

\end{remark}

\subsection{Multiplicative vs. additive coarse correction}
We are going to present the idea of two-level KFAC in a very elementary fashion. Let $m\geq \ell$ be an integer and $R_0\in \mathbb{R}^{m\times p}$ be a given matrix. The subspace of $\mathbb{R}^p$ spanned by the columns of $R_0^T \in \mathbb{R}^{p\times m}$ is called the \emph{coarse space}. The choice of the coarse space will be discussed later on. For the moment, we can assume that it is known.

The idea is to add to $\neginc_{\text{KFAC}}$ a correction term that lives in the coarse space, in such a way that the new vector minimizes the error in the $F_{\bullet}$-norm with respect to the FIM solution $\neginc = F_{\bullet}^{-1} \nabla_\theta h$. More concretely, this means that for the negative increment, we consider

\begin{equation}\label{eq:corrector}
\neginc_{\text{KFAC-2L}} = \neginc_{\text{KFAC}} + R_{0}^T\beta^{*},
\end{equation}
where
\begin{subequations}\label{eq:beta}
\begin{align} 
        \beta^{*} &= \argmin_{\beta \in \mathbb{R}^{\ell}}\| (\neginc_{\text{KFAC}}+R_{0}^T\beta)-\neginc \|_{F_{\bullet}}^2 \\
         &= \argmin_{\beta \in \mathbb{R}^{\ell}}\| (\neginc_{\text{KFAC}}+R_{0}^T\beta)- F_{\bullet}^{-1} \nabla_\theta h \|_{F_{\bullet}}^2 .
\end{align}
\end{subequations}
The solution of the quadratic minimization problem \eqref{eq:beta} is given by
\begin{equation}\label{eq:solminbeta}
\beta^{*} = (R_{0}F_{\bullet}R_{0}^T)^{-1}R_0 (\nabla_{\theta}h - F_{\bullet}\neginc_{\text{KFAC}}),
\end{equation}
provided that the matrix
\begin{equation}\label{eq:Fcoarse}
F_{\text{coarse}} := R_{0}F_{\bullet}R_{0}^T \in \mathbb{R}^{m \times m},
\end{equation}
representing the \emph{coarse operator}, be invertible. This is a small size matrix, insofar as $m$ will be in practice taken to be equal to $\ell$ or $2\ell$, and will in any case remain much smaller than $p$. This is in agreement with domain decomposition where the size of the coarse system is usually equal to the number of subdomains.

As for the vector
\begin{equation}\label{eq:residual}
r_{\text{KFAC}} := \nabla_{\theta}h - F_{\bullet}\neginc_{\text{KFAC}} ,
\end{equation}
it is referred to as the \emph{residual} associated to the approximate solution $\neginc_{\text{KFAC}}$.
Plugging \eqref{eq:solminbeta} into \eqref{eq:corrector} and recalling that $\neginc_{\text{KFAC}} = F_{\bullet\,\text{KFAC}}^{-1} \nabla_\theta h$, we end up with

\begin{equation}\label{eq:solution2L}
\neginc_{\text{KFAC-2L}} = F_{\bullet\,\text{KFAC-2L}}^{-1}\nabla_{\theta}h,
\end{equation}
with
\begin{equation}\label{eq:KFAC2LDD}
F_{\bullet\,\text{KFAC-2L}}^{-1} = F_{\bullet\,\text{KFAC}}^{-1} + R_0^T F_{\text{coarse}}^{-1} R_0 (I - F_{\bullet\phantom{K}}^{\phantom{-1}}\!\!\!\! F_{\bullet\,\text{KFAC}}^{-1}).
\end{equation}

The matrix \eqref{eq:KFAC2LDD} that we propose can be checked to be consistent: if $F_{\bullet\,\text{KFAC}}^{-1}$ and $R_0^T F_{\text{coarse}}^{-1} R_0$ are both homogeneous to $F_{\bullet}^{-1}$, then $F_{\bullet\,\text{KFAC-2L}}^{-1}$ is homogenous to
\[
F_{\bullet}^{-1} + F_{\bullet}^{-1} - F_{\bullet}^{-1} F_{\bullet}^{\phantom{-1}} \!\!\!\! F_{\bullet}^{-1} = F_{\bullet}^{-1}
\]
too. In the language of domain decomposition, the coarse corrector of \eqref{eq:KFAC2LDD} is said to act \emph{multiplicatively}, to the extent that
\begin{equation}\label{eq:multiplicative}
I - F_{\bullet\,\text{KFAC-2L}}^{-1} F_{\bullet\phantom{K}}^{\phantom{-1}} \!\!\!\! = [I - (R_0^TF_{\text{coarse}}^{-1}R_0)  F_{\bullet}] [I - F_{\bullet\,\text{KFAC}}^{-1} F_{\bullet\phantom{K}}^{\phantom{-1}} \!\!\!\!].
\end{equation}
as can be straightforwarldy verified. If $G$ is an approximation of $F_{\bullet}^{-1}$, the matrix $I- G F_{\bullet}$ measures the quality of this approximation. Equality \eqref{eq:multiplicative} shows that the approximation quality of $F_{\bullet\,\text{KFAC-2L}}^{-1}$ is the product of those of $R_0^TF_{\text{coarse}}^{-1}R_0$ and $F_{\bullet\,\text{KFAC}}^{-1}$.

A common practice in domain decomposition is to drop the factor $I - F_{\bullet\phantom{K}}^{\phantom{-1}} \!\!\!\! F_{\bullet\, \text{KFAC}}^{-1}$ (which is equivalent to replacing the residual $r_{\text{KFAC}} =\nabla_{\theta}h - F_{\bullet}\theta_{\text{KFAC}}$ by $\nabla_{\theta}h$). This amounts to approximating $F_{\bullet\,\text{KFAC-2L}}^{-1}$ as 
\begin{equation}\label{eq:bad2L}
F_{\bullet\,\text{KFAC-2L}}^{-1}\approx F_{\bullet\,\text{KFAC}}^{-1}+R_0^TF_{\text{coarse}}^{-1}R_0.
\end{equation}
The coarse corrector of \eqref{eq:bad2L} is said to act \emph{additively} in domain decomposition. Clearly, the resulting matrix is inconsistent with $F_{\bullet}^{-1}$: in fact, it is consistent with $2F_{\bullet}^{-1}$! No matter how crude it is, this coarse corrector is actually valid as long as
$F_{\bullet\,\text{KFAC-2L}}^{-1}$ is used only as a preconditioner in the resolution of the system $F_{\bullet}\neginc = \nabla_\theta h$, which means that we solve instead $F_{\bullet\,\text{KFAC-2L}}^{-1} F_{\bullet\phantom{K}}^{\phantom{-1}} \!\!\!\! \neginc = F_{\bullet\,\text{KFAC-2L}}^{-1}\nabla_\theta h$ to benefit from a more favorable conditioning but the solution we seek remains the same. 

Here, in our problem, $F_{\bullet}^{-1}$ is directly approximated by $F_{\bullet\,\text{KFAC-2L}}^{-1}$  and therefore the inconsistent additive coarse corrector \eqref{eq:bad2L} is not acceptable. Note that Tselepidis et al. \cite{twolevels} adopted this additive coarse correction, in which $F_{\text{coarse}}$ is approximated as
\begin{subequations}
\begin{equation}
F_{\text{coarse}}\approx R_{0}\Bar{F}_{\bullet} R_{0}^T,
\end{equation}
where $\Bar{F}_{\bullet}$ is the block-diagonal matrix whose blocks $[\Bar{F}_{\bullet}]_{i,j}$ are given by
\begin{equation}
[\Bar{F}_{\bullet}]_{i,j} = \begin{cases}
\, \mathbb{E}[\Bar{a}_{i-1}\Bar{a}_{j-1}^T]\otimes \mathbb{E}[g_ig_j^T] & \;\text{if }\; i\neq j,\\
\, \mathbb{E}[A_i + \pi_i\lambda^{1/2} I_{A_i}] \otimes \mathbb{E}[G_i + \pi_i^{-1} \lambda^{1/2} I_{G_i}] & \; \text{if } \; i=j.
\end{cases}
\end{equation}
\end{subequations}
In this work, we focus to the consistent multiplicative coarse corrector \eqref{eq:KFAC2LDD} and also consider the exact value \eqref{eq:Fcoarse} for $F_{\text{coarse}}$.

\subsection{Choice of the coarse space $R_0^T$}
By the construction \eqref{eq:corrector}--\eqref{eq:beta}, we are guaranteed that 
\begin{equation}\label{eq:diminution}
\| \neginc_{\text{KFAC-2L}} - \neginc \|_{F}^2 \leq \| \neginc_{\text{KFAC}} - \neginc \|_{F}^2
\end{equation}
for any coarse space $R_0^T$, since the right-hand side corresponds to $\beta = 0$. The choice of $R_0^T$ is a compromise between having a small dimension $m \ll p$ and lowering the new error
\begin{equation}
\big\| \neginc_{\text{KFAC-2L}} - \neginc \big\|_{F}^2 = \big\| - [I -  R_0^T (R_0 F_{\bullet} R_0^T)^{-1} R_0 F_{\bullet}][I- F_{\bullet\,\text{KFAC}}^{-1} F_{\bullet\phantom{K}}^{\phantom{-1}} \!\!\!\! ] \neginc \, \big\|_F^2
\end{equation}
as much as possible. But it seems out of reach to carry out the minimization of the latter with respect to the entries of $R_0^T$.

In the context of the preconditioner, the idea behind a two-level method is to remove first the influence of very large eigenvalues which correspond to high-frequency modes, then remove the smallest eigenvalues thanks to the second level, which affect greatly the convergence.
To do so, we need a suitable coarse space to efficiently deal with this second level \cite{Nataf2011}. Ideally, we would like to choose the deflation subspace which consists of the eigenvectors associated with the small eigenvalues of the preconditioned operator. However, this computation is more costly than solving a linear system itself.

This leads us to choose the coarse space in an a priori way. We consider the a priori form
\begin{equation}\label{eq:formeR0T}
R_{0}^T = \begin{bmatrix} 
    V_1 & 0 & \dots & \dots & 0 \\
     0 & V_2 & \dots & \dots & 0  \\
    \vdots & \vdots & \ddots &  &\vdots \\
    0 &  0  &  \dots  & \dots & V_{\ell} 
    \end{bmatrix} \in \mathbb{R}^{p\times m} ,
\end{equation}
where each block $V_i \in \mathbb{R}^{p_i\times m_i}$ has $m_i$ columns with $m_i \ll p_i$, and
\begin{equation}
m_1 + m_2 + \ldots + m_\ell = m .
\end{equation}
To provide a comparative study, we propose to evaluate several coarse space choices of the form \eqref{eq:formeR0T} that are discussed below.

\paragraph*{Nicolaides coarse space.} Historically, this is the first  \cite{Nicolaides1987DeflationOC} coarse space ever proposed in domain decomposition. Transposed to our case, it corresponds to
\begin{equation}\label{eq:m1lm}
m_1 = \ldots = m_\ell = 1, \qquad m = \ell,
\end{equation}
and for all $i\in \{1,\hdots,\ell\}$,
\begin{equation}\label{eq:nicospace}
V_i= \begin{bmatrix}
           1,\hdots, 1
         \end{bmatrix}^T \, \in \, \mathbb{R}^{p_i}.
\end{equation}
Originally, the motivation for selecting the vector whose all components are equal to 1 is that it is the discrete version of a continuous constant field, which is the eigenvector associated with the eigenvalue 0 of the operator $-\nabla \cdot (\kappa \nabla)$ (boundary conditions being set aside). Inserting it into the coarse space helps the solver take care of the lowest frequency mode. In our problem, however, there is no reason for 0 to be an eigenvalue of $F$, nor for 1 to be an eigenvector if this is the case. Hence, there is no justification for the Nicolaides coarse space. Still, this choice remains convenient and practical. This is probably the reason why Tselepidis et al. \cite{twolevels} have opted for it.

\paragraph*{Spectral coarse space.} This is a slightly refined version of the Nicolaides coarse space. The idea is always to capture the lowest mode \cite{Nataf2011}, but since the lowest eigenvalue and eigenvector are not known in advance, we have to compute them. More specifically, we keep the values \eqref{eq:m1lm} for the column sizes within $R_0^T$, while prescribing
\begin{equation}\label{eq:spectralspace}
V_i= \text{eigenvector associated to the smallest eigenvalue of } [F_{\bullet\,\text{KFAC}}]_{i,i}
\end{equation}
for all $i\in \{1,\hdots,\ell\}$. In our case, an advantageous feature of this definition is
that the cost of computing the eigenvectors is ``amortized'' by that of the inverses of $[F_{\text{KFAC}}]_{i,i}$, in the sense that these two calculations can be carried out simultaneously. Indeed, let
\begin{equation}
A_i + \pi_i\lambda^{1/2} I_{A_i} = U_{A_i}\Sigma_{A_i} V_{A_i}^T, \qquad 
G_i + \pi_i^{-1} \lambda^{1/2} I_{G_i} = U_{G_i}\Sigma_{G_i} V_{G_i}^T
\end{equation}
be the singular value decompositions of $A_i + \pi_i\lambda^{1/2} I_{A_i}$ and $G_i  + \pi_i^{-1} \lambda^{1/2} I_{G_i}$ respectively. Then,
\begin{align}
[F_{\bullet\,\text{KFAC}}]_{i,i}^{-1} &= (A_i + \pi_i\lambda^{1/2} I_{A_i} )^{-1}\otimes (G_i + \pi_i^{-1} \lambda^{1/2} I_{G_i})^{-1} \nonumber \\
  &= (U_{A_i}\Sigma_{A_i}V_{A_i}^T)^{-1}\otimes  (U_{G_i}\Sigma_{G_i}V_{G_i}^T)^{-1} \nonumber\\
  &= (U_{A_i}\Sigma_{A_i}^{-1}V_{A_i}^T)\otimes (U_{G_i}\Sigma_{G_i}^{-1}V_{G_i}^T).
\end{align}
Since $\Sigma_{A_i}$ and $\Sigma_{G_i}$ are diagonal matrices, their inverses are easy to compute. Now, if $V_{A_i}$ and $V_{G_i}$ are the eigenvectors associated to the smallest eigenvalues of $A_i$ and $G_i$ respectively, then the eigenvector associated to the smallest eigenvalue of $[F_{\bullet\,\text{KFAC}}]_{i,i}$ is given by
\begin{equation}
V_i = V_{A_i} \otimes V_{G_i}.
\end{equation}

\paragraph*{Krylov coarse space.} If we do not wish to compute the eigenvector associated to the smallest eigenvalue of $[F_{\bullet\,\text{KFAC}}]_{i,i}$, then a variant of the spectral coarse space could be the following. We know that this eigenvector can be obtained by the inverse power method. The idea is then to perform a few iterations of this method, even barely one or two, and to include the iterates into the the coarse subspace. If $m_i -1 \geq 1$ is the number of inverse power iterations performed for $[F_{\bullet\,\text{KFAC}}]_{i,i}$, then we take
\begin{equation}
    V_i = [v_i, \;\; [F_{\bullet\,\text{KFAC}}]_{i,i}^{-1} v_i, \;\; \ldots, \;\; [F_{\bullet\,\text{KFAC}}]_{i,i}^{-(m_i-1)} v_i] \,\in\, \mathbb{R}^{p_i\times m_i}
\end{equation}
where $v_i\in \mathbb{R}^{p_i}$ is an arbitrary vector, assumed to not be an eigenvector of $[F_{\bullet\,\text{KFAC}}]_{i,i}$ to ensure that the columns of $V_i$ are not collinear. By appropriately selecting $v_i$, we are in a position to use this approach to enrich the Nicolaides coarse space and the residuals coarse space (cf. next construction). 

The increase in the number of columns for $V_i$ is not the price to be paid to avoid the eigenvector calculation: we could have put only the last iterate $[F_{\bullet\, \text{KFAC}}]_{i,i}^{-(m_i-1)} v_i$ into $V_i$. But since we have computed the previous ones, it seems more cost-effective to use them all to enlarge the coarse space. The larger the latter is, the lower is the minimum value of the objective function. In this work, we consider the simplest case

\begin{equation}\label{eq:m2l}
    m_1 = \ldots = m_\ell = 2, \qquad m = 2\ell.
\end{equation}

\paragraph*{Residuals coarse space.} We now introduce a very different philosophy of coarse space, which to our knowledge has never been envisioned before. 
From the construction \eqref{eq:corrector}--\eqref{eq:beta}, it is obvious that if the error $\neginc - \neginc_{\text{KFAC}}$ belongs to the coarse space $R_0^T$, that is, if it can be written as a linear combination $R_0^T \beta^\sharp$ of the coarse matrix columns, then the vector $\neginc_{\text{KFAC}} + R_0^T\beta^{\sharp}$ coincides with the exact solution $\neginc$ and the correction would be ideally optimal. Although this error $\neginc - \neginc_{\text{KFAC}}$ is unknown, it is connected to the residual \eqref{eq:residual} by
\begin{equation}\label{eq:errFres}
\neginc - \neginc_{\text{KFAC}} = F_{\bullet}^{-1}r_{\text{KFAC}} .
\end{equation}
The residual $r_{\text{KFAC}}$ is not too expensive to compute. as it consists of a direct matrix-product $F\neginc_{\text{KFAC}}$.
Unfortunately, solving a linear system involving $F$ as required by \eqref{eq:errFres} is what we want to avoid.

But we can just approximate this error by inverting with $F_{\bullet\,\text{KFAC}}^{-1}$ instead of $F_{\bullet}^{-1}$. Therefore, we propose to build a coarse space that contains $F_{\bullet\,\text{KFAC}}^{-1}r_{\text{KFAC}}^{\phantom{-1}}$ instead of $F_{\bullet}^{-1}r_{\text{KFAC}}$. To this end, we split $F_{\bullet\, \text{KFAC}}^{-1}r_{\text{KFAC}}$ into $\ell$ segments, each corresponding to a layer. This amounts to choosing the values \eqref{eq:m1lm} for the column sizes and set the columns of $R_0^T$ as
\begin{equation}\label{eq:residuspace}
V_i =[F_{\bullet\,\text{KFAC}}]_{i,i}^{-1}r_{\text{KFAC}}[i] \in \mathbb{R}^{p_i}, \qquad r_{\text{KFAC}}[i] = \vecteur(\mathcal{D}W_i)- ( F_{\bullet}\neginc_{\text{KFAC}})[i]
\end{equation}
for $i\in\{1,\ldots,\ell\}$, where for a vector $\xi\in\mathbb{R}^p$ the notation $\xi[i] = \xi(p_{i-1}+1:p_i)$ designates the portion related to layer $i$. Formulas \eqref{eq:residuspace} ensure that $F_{\bullet\,\text{KFAC}}^{-1}r_{\text{KFAC}}$ belongs to the coarse space. Indeed, taking $\beta =[1,\hdots,1]^T \in \mathbb{R}^{\ell}$, we find $R_0^T\beta =F_{\bullet\,\text{KFAC}}^{-1}r_{\text{KFAC}}^{\phantom{-1}}$. 

\paragraph*{Taylor coarse space.} The previous coarse space is the zeroth-order representative of a family of more sophisticated constructions based on a formal Taylor expansion of $F_{\bullet}^{-1}$, which we now present but which will not be implemented. Setting
\begin{equation}\label{eq:qualitymatrix}
E = I - F_{\bullet\,\text{KFAC}}^{-1} F_{\bullet\phantom{K}}^{\phantom{-1}} \!\! 
\end{equation}
and observing that $F_{\bullet} = F_{\bullet\,\text{KFAC}} (I- E)$, we have
\begin{equation}
F_{\bullet}^{-1} = (I- E)^{-1} F_{\bullet\,\text{KFAC}}^{-1} = (I+ E + \ldots + E^{q-1}+ \ldots) F_{\bullet\,\text{KFAC}}^{-1}.
\end{equation}
The formal series expansion in the last equality rests upon the intuition that $E$ measures the approximation quality of $F_{\bullet}^{-1}$ by $F_{\bullet\,\text{KFAC}}^{-1}$ and therefore can be assumed to be small. Multiplying both sides by the residual $r_{\text{KFAC}}$ and stopping the expansion at order $q-1\geq 0$, we obtain the approximation
\begin{equation}\label{eq:taylorcorr}
(I+ E + \ldots + E^{q-1}) F_{\bullet\,\text{KFAC}}^{-1} r_{\text{KFAC}}^{\phantom{-1}}
\end{equation}
for the error $F_{\bullet}^{-1} r_{\text{KFAC}} = \neginc - \neginc_{\text{FAC}}$, which is also the ideal correction term. As earlier, we impose that this approximate correction vector \eqref{eq:taylorcorr} must be contained in the coarse space $R_0^T$. This suggests to extract the components in layer $i$ of the vectors
\[
\big\{ F_{\bullet\,\text{KFAC}}^{-1} r_{\text{KFAC}}^{\phantom{-1}}, \; E F_{\bullet\,\text{KFAC}}^{-1} r_{\text{KFAC}}^{\phantom{-1}}, \;\ldots, \; E^{q-1} F_{\bullet\,\text{KFAC}}^{-1} r_{\text{KFAC}}^{\phantom{-1}} \big\}
\]
and assign them to the columns of $V_i$. In view of \eqref{eq:qualitymatrix}, the space spanned by the above vectors is the same as the one spanned by
\[
\big\{ F_{\bullet\,\text{KFAC}}^{-1} r_{\text{KFAC}}^{\phantom{-1}}, \; (F_{\bullet\, \text{KFAC}}^{-1} F_{\bullet\phantom{K}}^{\phantom{-1}} \!\!\!\! ) F_{\bullet\,\text{KFAC}}^{-1} r_{\text{KFAC}}^{\phantom{-1}}, \;\ldots, \; (F_{\bullet\,\text{KFAC}}^{-1} F_{\bullet\phantom{K}}^{\phantom{-1}}\!\!\!\! )^{q-1} F_{\bullet\,\text{KFAC}}^{-1} r_{\text{KFAC}}^{\phantom{-1}} \big\}.
\]
Consequently, we can take
\begin{equation}
    m_1 = \ldots = m_\ell = q, \qquad m = q\ell,
\end{equation}
and
\begin{equation}
    V_i = [ w_1[i], \; w_2[i], \; \ldots , \; w_q[i] ] \in \mathbb{R}^{p_i\times m_i}
\end{equation}
where
\begin{equation}\label{eq:mycolumns}
w_1 = F_{\bullet\,\text{KFAC}}^{-1} r_{\text{KFAC}}^{\phantom{-1}} \in \mathbb{R}^{p}, \qquad w_{j+1} = F_{\bullet\,\text{KFAC}}^{-1} F_{\bullet\phantom{K}}^{\phantom{-1}} \!\!\!\! w_j \in \mathbb{R}^{p},
\end{equation}
for $1\leq j\leq q-1$. The case $q=1$ degenerates to the residuals coarse space. From \eqref{eq:mycolumns}, we see that upgrading to the next order is done by multiplying by $F_{\bullet}$, an operation that mixes the layers.

For the practical implementation of these coarse spaces, we need efficient computational methods for two essential building blocks, namely, the matrix-vector product $F_{\bullet} u$ and the coarse operator $F_{\text{coarse}}$. These will be described in appendix \S\ref{appendix}.

%\subsection{Computation of the coarse operator}\label{sec:computations}

\subsection{Pseudo-code for two-level KFAC methods}
Algorithm \ref{algo} summarizes the steps for setting up a two-level KFAC method. 

\begin{algorithm}[H]\label{algo}
\SetAlgoLined
\KwInput{$\theta_0$ (Initial point), $k_{\max}$ (maximum number of iterations), and $\alpha$ (learning rate)}
\KwOutput{$\theta_{k_{\max}}$}
 \For{$k=0,1,\hdots,k_{\max}-1$}{
 %$\bullet$ Select a mini-batch $\mathcal{B} \subset \mathcal{U}$\;
 $\bullet$ Compute an estimate $\nabla_{\theta}h({\mathcal{S}_k,\theta_{k}})$ of the gradient on a mini-batch $\mathcal{S}_k$ randomly sampled from the training data\;
$\bullet$ Compute $\theta_{\text{KFAC}}=F_{\bullet\,\text{KFAC}}^{-1} \nabla_{\theta}h({\mathcal{S}_k,\theta_{k}})$\;
$\bullet$ Choose a coarse space $R_0^T$ and compute the associated coarse correction $R_0\beta^{*}=R_{0}^T(F_{\text{coarse}})^{-1}R_{0}r_{\text{KFAC}}$\;
$\bullet$ Compute $\theta_{\text{KFAC-2L}}=\theta_{\text{KFAC}}+R_0\beta^{*}$\;
$\bullet$ Update $\theta_{k+1}=\theta_k-\alpha \theta_{\text{KFAC-2L}}$\;
 }
 \caption{ High-level pseudo-code for a two-level KFAC method}
\end{algorithm}

\section{Numerical results}\label{sec:experiments}
In this section, we 
compare the new two-level KFAC methods designed in \S\ref{sec:Two-level} with the standard KFAC \cite{MartensGrosse2015,GrosseMartens2016} from the standpoint of convergence speed. For a thorough analysis, we also include the two-level KFAC version of Tselepidis et al. \cite{twolevels} and baseline optimizers (ADAM and SGD).

We run a series of experiments to investigate the optimization performance of deep auto-encoders, CNNs, and deep linear networks. Since our primary focus is on convergence speed rather than generalization, we shall only be concerned with the ability of optimizers to minimize the objective function. In particular, we report only training losses for each optimizer. To equally treat all methods, we adopt the following rules.
We perform a Grid Search and select hyper-parameters that give the best reduction to the training loss. Learning rates for all methods and damping parameters for KFAC and two-level KFAC methods are searched in the range
\[
\{10^{-4}, \, 10^{-3}, \, 10^{-2}, \, 10^{-1}, \, 10^{0}, \, 10^{1}, \, 10^{2},  \, 10^{3}, \, 10^{4}\}.
\]
For each optimizer, we apply the Early Stopping technique with patience of 10 epochs i.e.  we stop training the network when there is no decrease in the training loss during 10 consecutive epochs). We also include weight decay with a coefficient of $10^{-3}$ for all optimizers.

All experiments presented in this work are performed with PyTorch framework \cite{pytorch} on a supercomputer with Nvidia Ampere A100 GPU and AMD Milan@2.45GHz CPU. 
For ease of reading, the following table explains all abbreviations of two-level KFAC methods that we will use in the figure legends.

\begin{table}[htb]
\tbl{Name abbreviations of two-level KFAC optimizers.}
{\begin{tabular}{ll} \toprule
 Optimizer & Name abbreviation \\ \midrule
 Two-level KFAC with Nicolaides coarse space  & NICO \\
 Two-level KFAC with spectral coarse space  & SPECTRAL \\
 Two-level KFAC with residuals coarse space & RESIDU\\
 Two-level KFAC with Krylov Nicolaides coarse space & KRY-NICO\\
Two-level KFAC with Krylov residuals coarse space & KRY-RESIDU\\
Two-level KFAC of Tselepidis et al. \cite{twolevels} & PREVIOUS \\
 \bottomrule
\end{tabular}}

\label{sample-table}
\end{table}

\subsection{Deep auto-encoder problems}
The first set of experimental tests performed is the optimization of three different deep auto-encoders, each trained with a different dataset (CURVES, MNIST, and FACES). Note that due to the difficulty of optimizing the underlying networks, these three auto-encoder problems are commonly used as benchmarks for evaluating new optimization methods in the deep learning community \cite{martens2010,sluskeretal2013,MartensGrosse2015,BRB2017,Koroko-et-al}. For each problem, we train the network with three different batch sizes. 

\begin{figure*}[h]
\centering
\begin{tabular}{c}
      \includegraphics[width=1.0\textwidth]{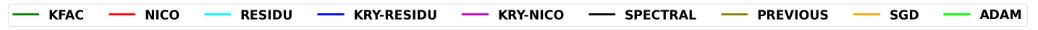} \\
      \subfloat[CURVES]{\includegraphics[width=140mm]{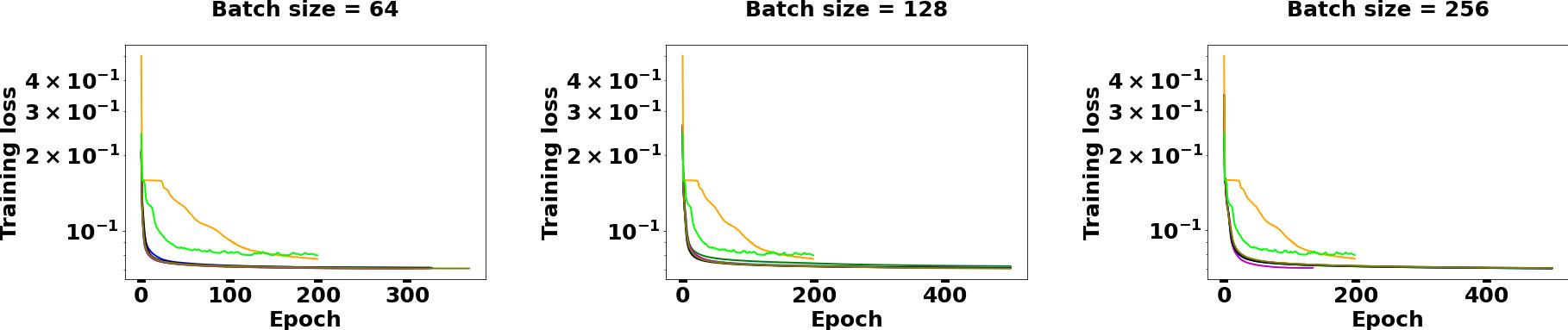}} \\
      \subfloat[MNIST]{\includegraphics[width=140mm]{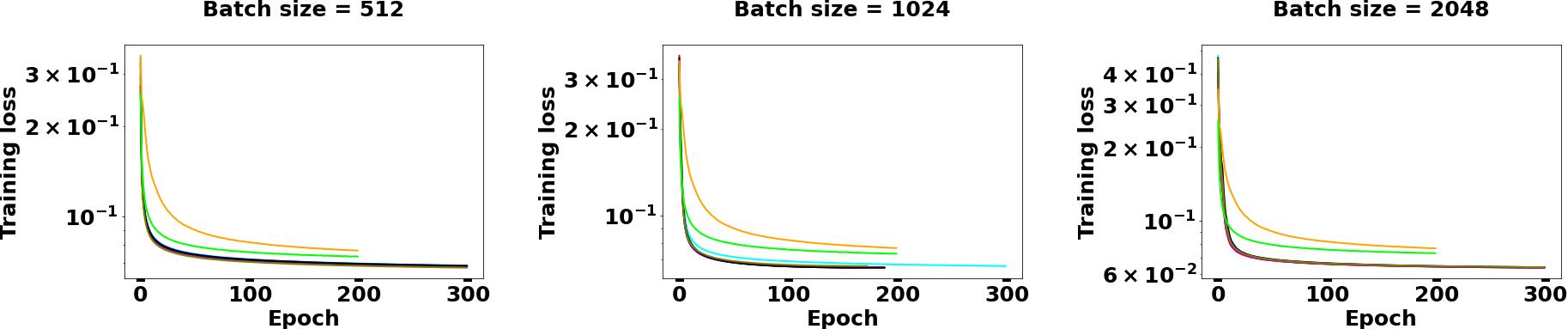}}\\      
     \hspace{0.1cm}\subfloat[FACES]{\includegraphics[width=137mm]{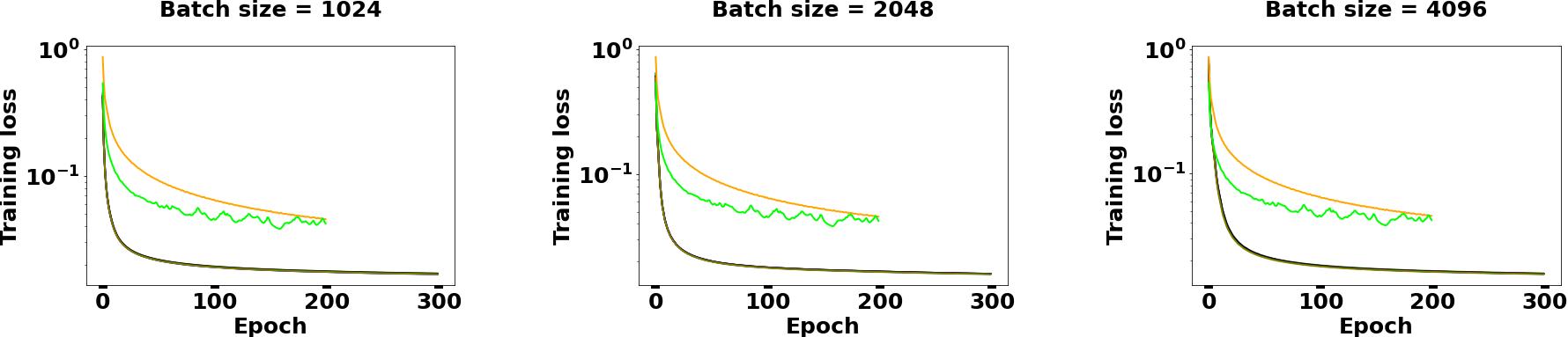}}
\end{tabular}
\caption{Comparison of KFAC against two-level KFAC methods on the three deep auto-encoder problems (CURVES \textbf{top} row, MNIST \textbf{middle} row and FACES \textbf{bottom} row). Three different batch sizes are considered for each problem (each column corresponds to a different batch size).}
\label{fig:autoencoder}
\end{figure*}

Figure \ref{fig:autoencoder} shows the obtained results. The first observation is that, as expected, natural gradient-based methods (KFAC and two-level KFAC methods) outperform baseline optimizers (ADAM and SGD). The second and most important observation is that, for each of the three problems, regardless of the batch size, the training curve of KFAC and those of all two-level KFAC methods (the one of Tselepidis et al. \cite{twolevels} and those proposed in this work) are overlaid, which means that taking into account the extra-diagonal terms of the Fisher matrix through two-level decomposition methods does not improve the convergence speed of KFAC method. This second observation is quite puzzling, since theoretically  two-level methods are supposed to offer a better approximation to the exact natural gradient than KFAC does and therefore should at least slightly outperform KFAC in terms of optimization performance. Note that we repeated these experiments on three different random seeds and obtained very similar results.

These surprising results are in line with the findings of Benzing \cite{Benzing2022GradientDO}, according to which KFAC outperforms the exact natural gradient in terms of optimization performance. This suggests that extra-diagonal blocks of the FIM do not contribute to improving the optimization performance, and sometimes even affect it negatively.

\subsection{Convolution neural networks}

The second set of experiments concerns the optimization of three different CNNs namely Resnet 18 \cite{HeEtal2016}, Cuda-convnet and Resnet 34 \cite{HeEtal2016}. We consider in particular Cuda-convnet which is the architecture used to evaluate the original KFAC method in \cite{GrosseMartens2016}. It must be mentioned that it contains $3$ convolution layers and one MLP layer. We train Cuda-convnet on CIFAR10 dataset \cite{Krizhevsky2009LearningML} with a batch size equal to $256$, and Resnet 18 on CIFAR100 \cite{Krizhevsky2009LearningML} with a batch size equal to $128$. Finally, we train Resnet 34 on the SVHN dataset \cite{Netzer2011ReadingDI} with a batch size equal to~$512$. 

For these CNNs (see Figure \ref{fig:CNN}), we arrive at  quite similar observations and conclusions to those we mention for  deep auto-encoder problems. In particular, like in \cite{twolevels}, when considering CNNs, we do not observe any significant gain in the convergence speed of KFAC when we enrich it with cross-layer information through two-level decomposition methods. Once again, these results corroborate the claims of Benzing \cite{Benzing2022GradientDO} and suggest that we do not need to take into account the extra diagonal blocks of the FIM.

\begin{figure*}[thbp]
\centering
\begin{tabular}{c}
      \includegraphics[width=1.0\textwidth]{Figures/Legend.JPG} \\
       \includegraphics[width=142mm]{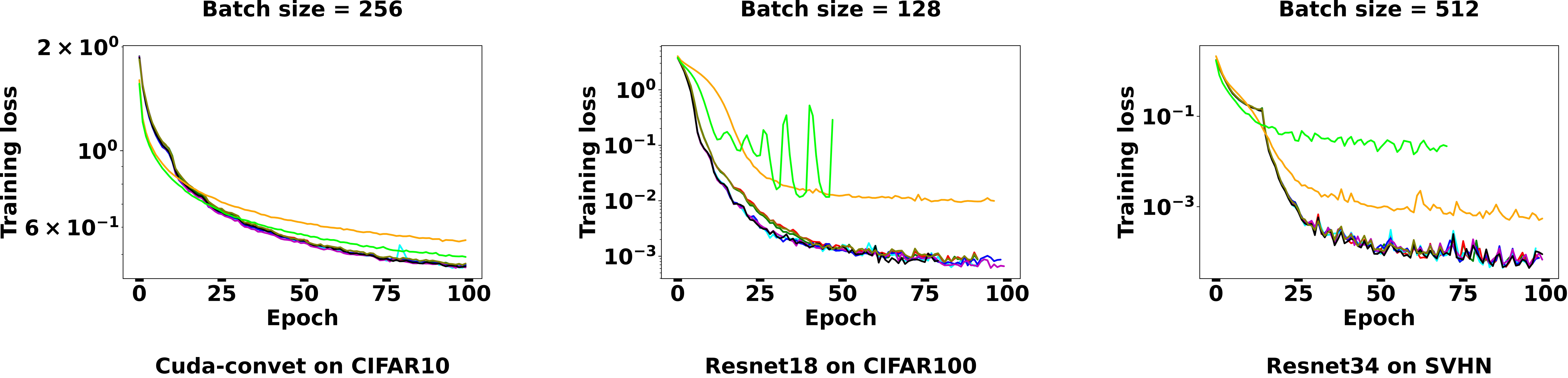} \\
\end{tabular}
\caption{Optimization performance evaluation of KFAC and two-level KFAC methods on three different CNNs.}
\label{fig:CNN}
\end{figure*}
\subsection{Deep linear networks}\label{subsec:linear}
The last experiments concern relatively simple optimization problems: linear networks optimization. 
We consider two deep linear networks.  These tests are motivated by the results obtained by Tselepidis et al. \cite{twolevels} for their two-level method. Indeed, for an extremely simple linear network with 64 layers (each
layer contains 10 neurons and a batch normalization layer) trained with randomly generated ten-size input vectors, they outperform KFAC in terms of optimization performance. Here, we first consider the same architecture but train the network on the Fashion
MNIST dataset \cite{xiao2017/online}(since we could not use the
same dataset). Then, we consider another linear network that contains 14 layers with batch
normalization, with this time much larger layers. More precisely we consider the following architecture: $784-1000-900-800-700-600-500-400-300-200-100-50-20-10$. We train this second network on the MNIST dataset. Both networks are trained with a batch size of $512$.

\begin{figure}[h]
\centering
\includegraphics[width=1.0\textwidth]{Figures/Legend.JPG}
\subfloat[Deep linear network of 64 layers, each layer containing 10 neurons, trained on Fashion MNIST.]{%
\resizebox*{6.5cm}{!}{ \includegraphics{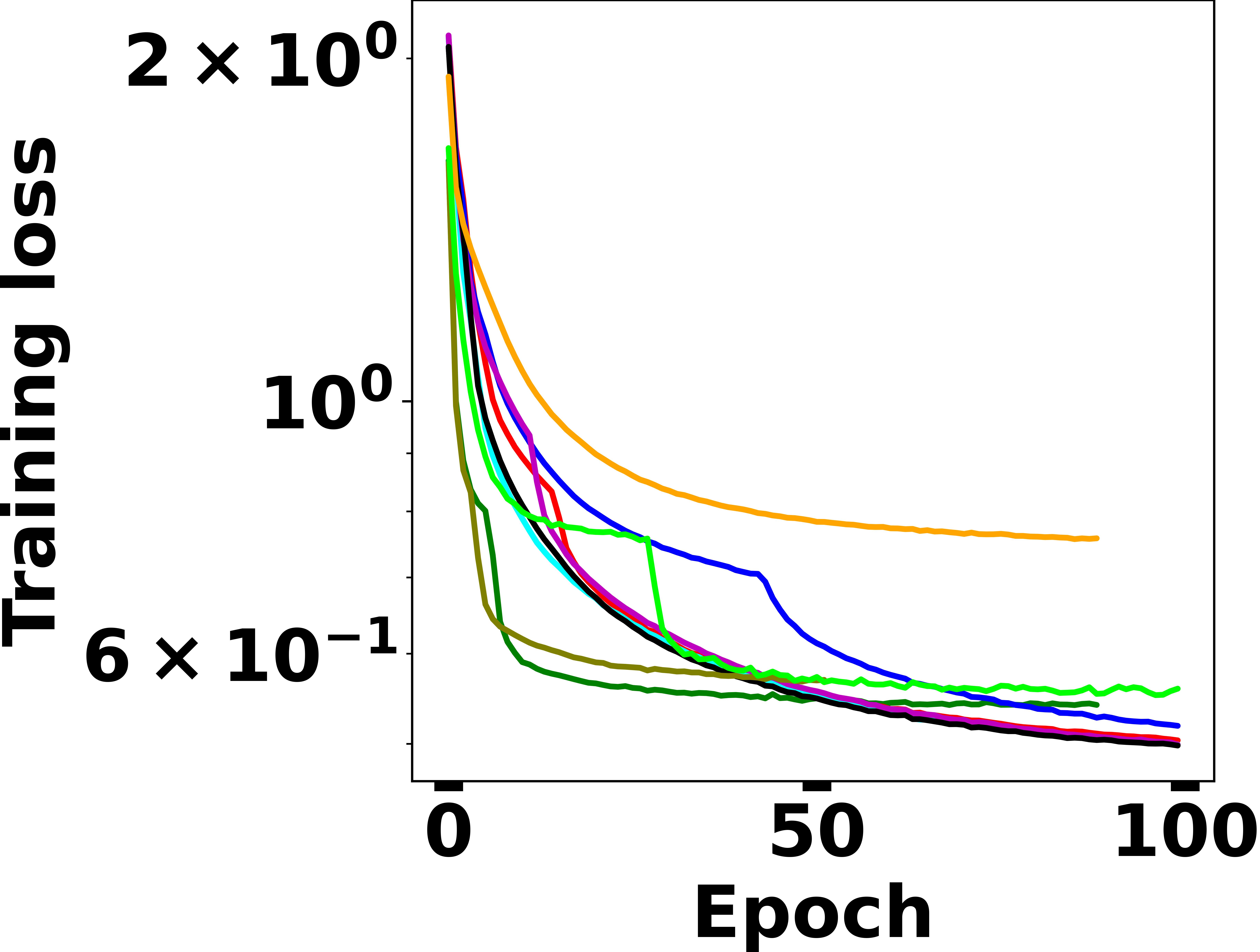}}}\hspace{8pt}
\subfloat[Deep linear network of 14 larger layers, trained on MNIST.]{%
\resizebox*{6.5cm}{!}{ \includegraphics{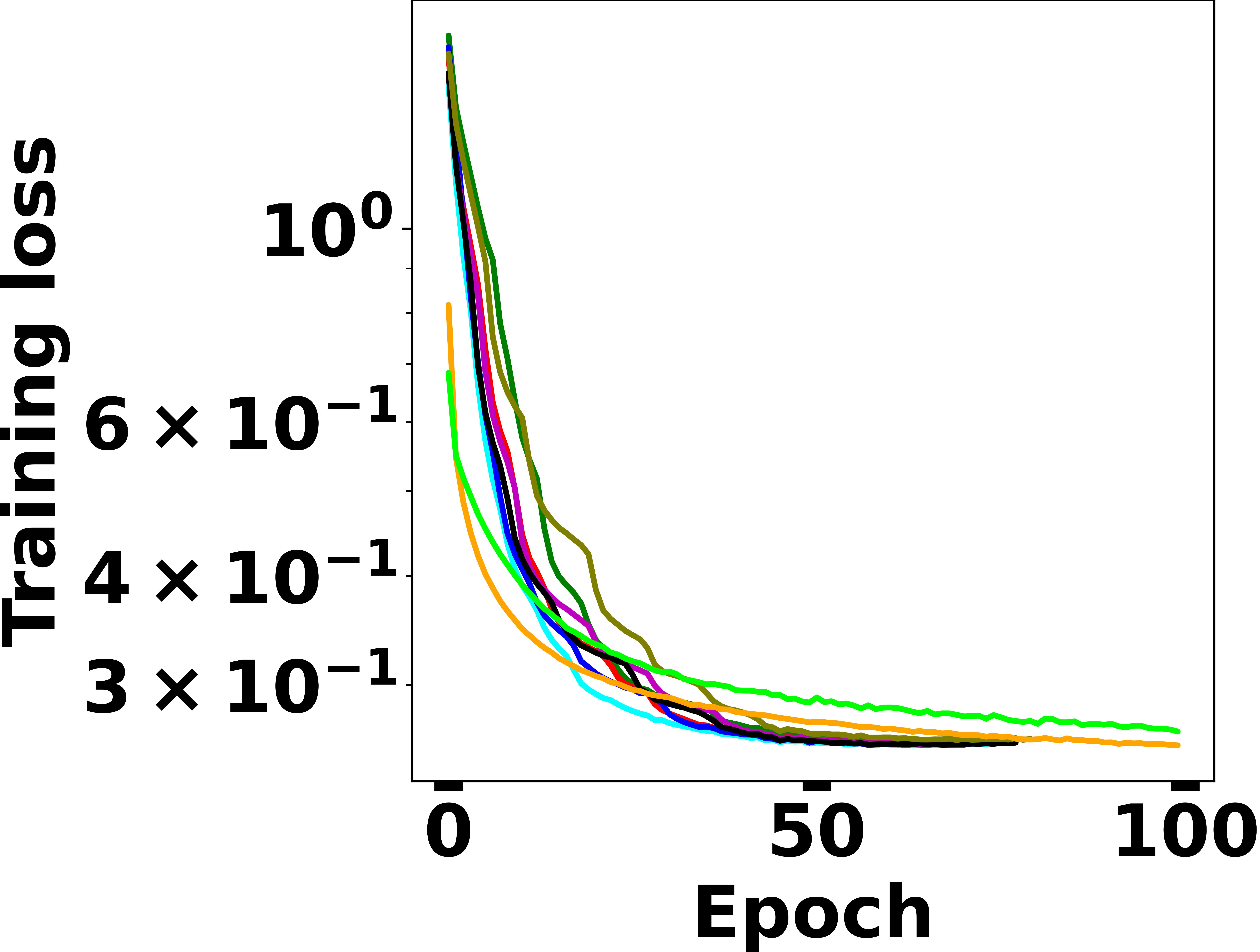}}}\hspace{10pt}
\caption{Optimization performance evaluation of KFAC and Tow-level KFAC optimizers on two different deep linear networks.} \label{fig:linear}
\end{figure}

Figure \ref{fig:linear} shows the training curves obtained in both cases. Here we observe like in \cite{twolevels} an improvement  in the optimization performance of two-level optimizers over KFAC. However, this gain remains too small and only concerns simple linear networks that are not used for practical applications. We therefore do not encourage enriching KFAC with two-level methods that require additional computational costs.

\subsection{Verification of error reduction for linear systems}

In the above experiments, two-level methods do not seem to outperform KFAC in terms of optimization performance. We thus wish to check that at each descent iteration, the negative increment $\neginc_{\text{KFAC-2L}}$ obtained with the coarse correction is indeed closer to that of the regularized natural gradient one $\neginc$ than the negative increment $\neginc_{\text{KFAC}}$ corresponding to the original KFAC. In other words, we want to make sure that inequality \eqref{eq:diminution} holds numerically.

For $\beta\in\mathbb{R}^m$, let
\begin{equation}
    \mathfrak{E}(\beta) = \| \neginc_{\text{KFAC}} + R_0^T \beta - \neginc \|_{F_{\bullet}}^2
\end{equation}
be the function to be minimized at fixed $R_0^T$ in the construction \eqref{eq:corrector}--\eqref{eq:beta}, where it is recalled that
\[
\neginc = F_{\bullet}^{-1} \nabla_\theta h, \qquad
\neginc_{\text{KFAC}} = F_{\bullet\,\text{KFAC}}^{-1} \nabla_\theta h .
\]
Note that
\begin{equation}
    \mathfrak{E}(0) = \| \neginc_{\text{KFAC}} - \neginc \|_{F_{\bullet}}^2
\end{equation}
is the squared $F_{\bullet}$-distance between the KFAC increment and that natural gradient one, regardless of $R_0^T$. Meanwhile, if $\beta^*$ is taken to be the optimal value \eqref{eq:solminbeta}, then
\begin{equation}
    \mathfrak{E}(\beta^*) = \| \neginc_{\text{KFAC-2L}} - \neginc \|_{F_{\bullet}}^2 .
\end{equation}
To see whether \eqref{eq:diminution} is satisfied, the idea is to compute the difference $\mathfrak{E}(\beta^*) - \mathfrak{E}(0)$ and check that it is negative. 
The goal of the game, however, is to avoid using the unknown natural gradient solution $\neginc$. Owing to the identity $\|a\|^2 - \|b\|^2 = (a-b,a+b)$ for the $F_{\bullet}$-dot product, this difference can be transformed into
\begin{align}
\mathfrak{E}(\beta^*) - \mathfrak{E}(0) &= \| \neginc_{\text{KFAC-2L}} - \neginc \|_{F_{\bullet}}^2 - \| \neginc_{\text{KFAC}} - \neginc \|_{F_{\bullet}}^2 \nonumber\\
&= (\neginc_{\text{KFAC-2L}} - \neginc_{\text{KFAC}}, \, \neginc_{\text{KFAC-2L}} + \neginc_{\text{KFAC}} - 2 \neginc)_{F_{\bullet}} \nonumber\\
&= \| \neginc_{\text{KFAC-2L}} - \neginc_{\text{KFAC}} \|_{F_{\bullet}}^2 + 2 (\neginc_{\text{KFAC-2L}} - \neginc_{\text{KFAC}}, \,  \neginc_{\text{KFAC}} - \neginc)_{F_{\bullet}} \nonumber\\
&= \| R_0^T \beta^* \|_{F_{\bullet}}^2 + 2( R_0^T \beta^*, \, \neginc_{\text{KFAC}} - \neginc)_{F_{\bullet}} \nonumber\\
&= \big \langle F_{\bullet} R_0^T \beta^* , R_0^T \beta^* \big\rangle + 2 \big\langle R_0^T \beta^*, \, F_{\bullet} (\neginc_{\text{KFAC}} - \neginc) \big\rangle ,
\end{align}
where $\langle \cdot,\cdot\rangle$ denotes the Euclidean dot product. But
\begin{equation}
    F_{\bullet} (\neginc_{\text{KFAC}} - \neginc) = F_{\bullet} \neginc_{\text{KFAC}} - \nabla_\theta h = - r_{\text{KFAC}}
\end{equation}
is the opposite of the residual \eqref{eq:residual}, which can be computed without knowing $\neginc$. Finally, the desired difference can also be computed as

\begin{equation}\label{eq:Ediff}
    \mathfrak{E}(\beta^*) - \mathfrak{E}(0) = \big\langle R_0 F_{\bullet} R_0^T \beta^*, \, \beta^* \big\rangle - 2 \big\langle R_0^T \beta^*, \, r_{\text{KFAC}} \big\rangle .
\end{equation}

For the two-level method of Tselepidis-Kohler-Orvieto \cite{twolevels}, the correction reads
\begin{subequations}
\begin{equation}
    \neginc_{\text{TKO}} = \neginc_{\text{KFAC}} + R_0^T \beta^*_{\text{TKO}}
\end{equation}
with
\begin{equation}
    \beta^*_{\text{TKO}} = (R_0 F_{\bullet} R_0^T)^{-1} R_0 \nabla_\theta h
\end{equation}
\end{subequations}
instead of $\beta^*$, the KFAC-2L value \eqref{eq:solminbeta}. The difference $\mathfrak{E}(\beta^*_{TKO}) - \mathfrak{E}(0)$ is then given by a formula similar to \eqref{eq:Ediff} in which $\beta^*$ is simply replaced by $\beta^*_{\text{TKO}}$.

\begin{figure*}[h]
\centering
\includegraphics[width=0.8\textwidth]{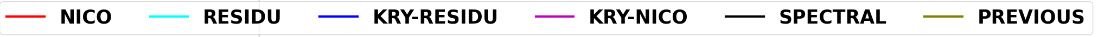}
\begin{tabular}{cc}
      \subfloat[CURVES, batch size $=256$]{ \includegraphics[width=60mm]{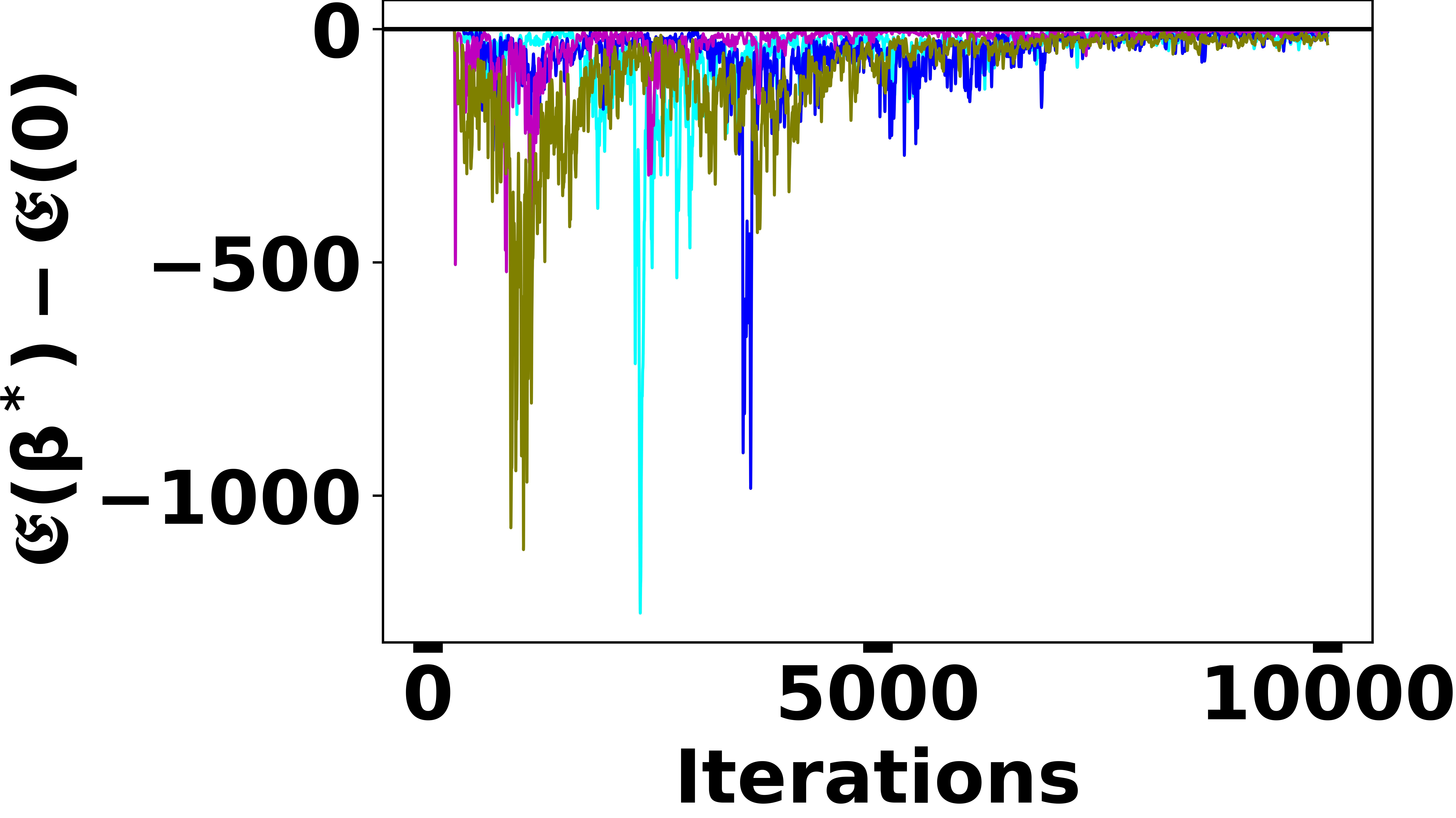}}  & 
      \subfloat[MNIST, batch size $=512$]{ \includegraphics[width=60mm]{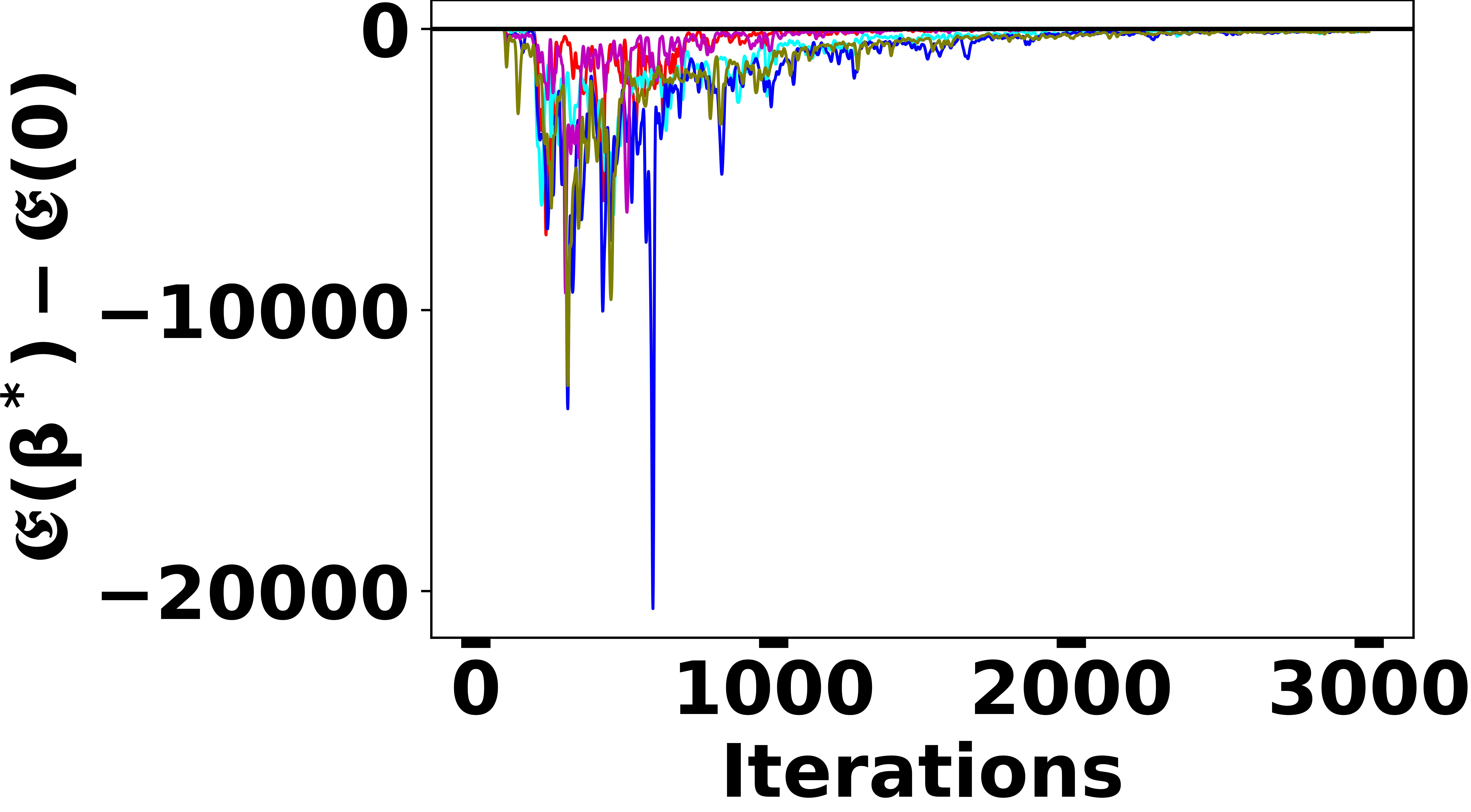}} \\
      \subfloat[FACES, batch size=$1024$]{ \includegraphics[width=56mm]{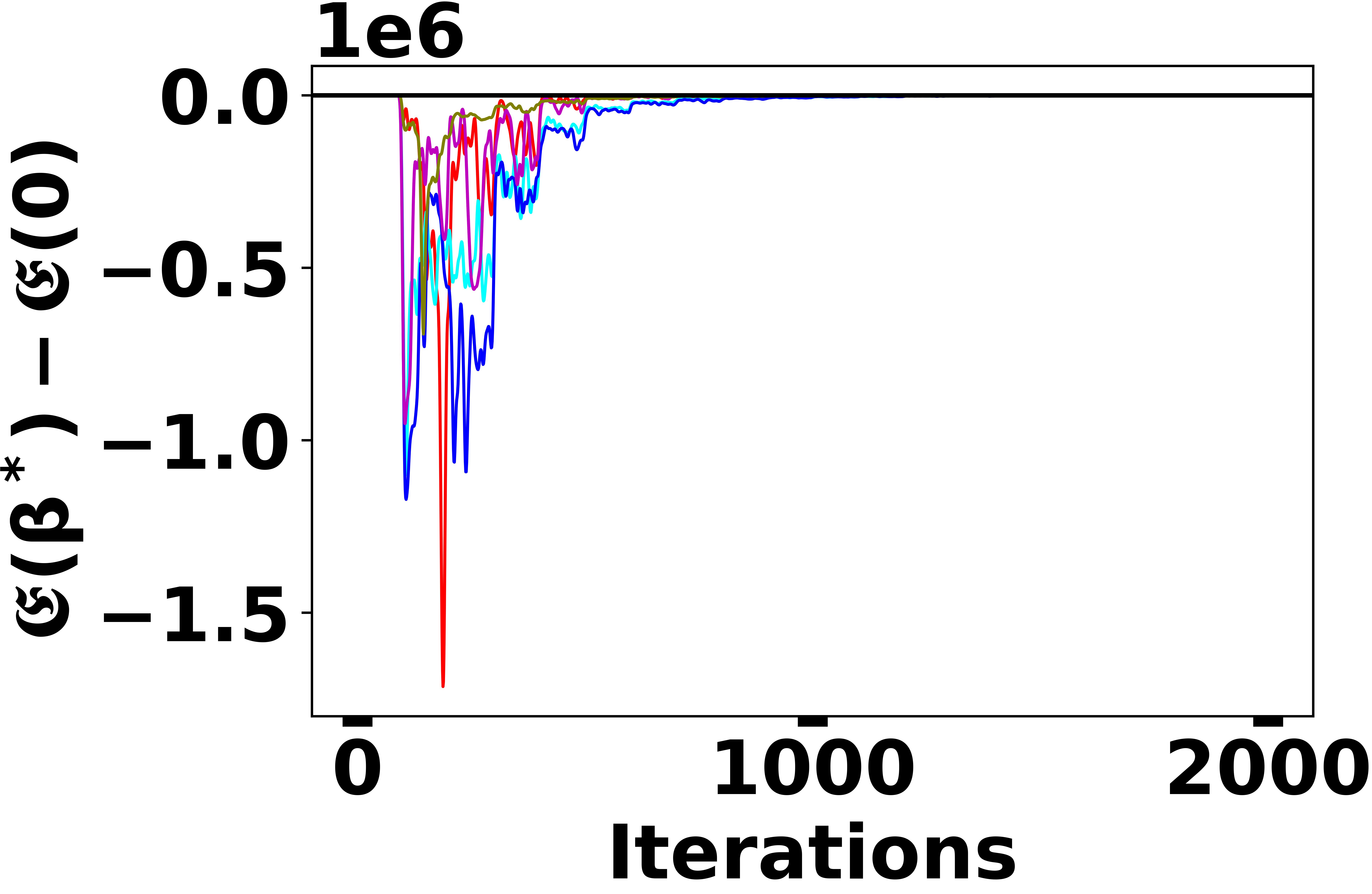}}  & \hspace{0.5cm}
      \subfloat[Cuda-convnet, batch size=$256$]{ \includegraphics[width=60mm]{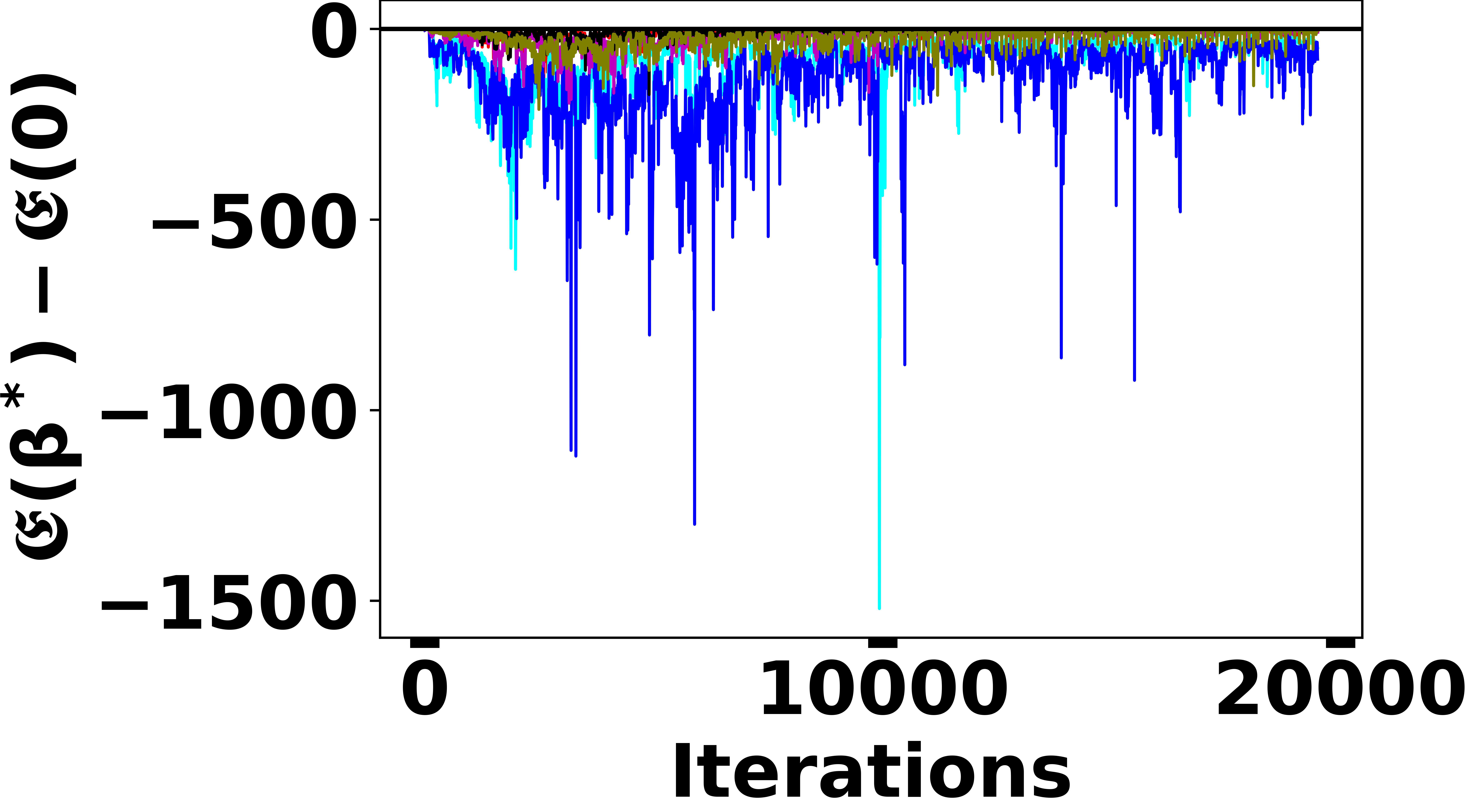}}
\end{tabular}
\caption{Evolution of $\mathfrak{E}(\beta^{*})-\mathfrak{E}(0)$ during training for each of the two-level methods considered. All methods proposed in this work as well as the TKO two-level method \cite{twolevels} have the gap $\mathfrak{E}(\beta^{*}) - \mathfrak{E}(0)$ negative throughout the training process.}
\label{fig:RESIDUS}
\end{figure*}

We compute the error $\mathfrak{E}(\beta^{*}) - \mathfrak{E}(0)$ associated to various two-level methods in the experiments conducted above. More specifically, we do it for the three deep auto-encoder problems and also for a CNN (cuda-convnet). The results obtained are shown in Figure \ref{fig:RESIDUS}.  The observation is that all two-level methods proposed in this work as well as the TKO two-level method \cite{twolevels} have the gap have negative gaps $\mathfrak{E}(\beta^{*})-\mathfrak{E}(0)$ throughout the optimization process. This implies that two-level methods solve the linear system \eqref{eq:sys} more accurately than KFAC does. It also means that the approximate natural gradients obtained with Two-level methods are closer to the exact natural gradient than the one obtained with KFAC is. 

\newpage
\section{Conclusion and discussion}\label{sec:conclusion}

In this study, we sought to improve KFAC by incorporating extra-diagonal blocks using two-level decomposition methods. To this end, we proposed several two-level KFAC methods, with a careful design of coarse corrections. Through several experiments, we came to the conclusion that two-level KFAC methods do not generally outperform the original KFAC method in terms of optimization performance of the objective function. This implies that taking into account the interactions between the layers is not useful for the optimization process.

We also numerically verfied that, at the level of the linear system of each iteration, the increment provided by any two-level method is much closer to the exact natural gradient solution than that obtained with KFAC, in a norm naturally associated with the FIM. This reveals that closeness to the exact natural gradient does not necessarily results in a more efficient algorithm. This observation is consistent with Benzing's previous claim \cite{Benzing2022GradientDO} that KFAC outperforms the exact natural gradient in terms of optimization performance. 

The fact that incorporating extra-diagonal blocks does not improve or often even hurts the optimization performance of the initial diagonal approximation could be explained by a negative interaction between different layers of the neural network. This suggests ignoring extra-diagonal blocks of the FIM and keeping the block-diagonal approximation, and if one seeks to improve the block-diagonal approximation, one should focus on diagonal blocks as attempted in many recent works \cite{Koroko-et-al,TKFAC,GLBBV2018,BRB2017}. 

It is worth pointing out that the conclusion of Tpselepedis et al. \cite{twolevels} on the performance of their proposed two-level method seems a little hasty.  Indeed, the authors only ran two different experiments: the optimization of a CNN and a simple linear network. For the CNN network, they did not observe any improvement. For the linear network they obtain some improvement in the optimization performance. Their conclusion is therefore based on this single observation.

Finally, we recall that as is the case for almost every previous work related to natural gradient and KFAC methods \cite{MartensGrosse2015,BRB2017,GrosseMartens2016,Benzing2022GradientDO}, the one undertaken in this paper is limited to the optimization performance of the objective function. It will thus be interesting to investigate the generalization capacity of these methods (including KFAC). Since the study of generalization requires a different experimental framework \cite{Benzing2022GradientDO,Zhang2019,ZangGrosse2018}, we leave it as a prospect. Our findings and those of Benzing \cite{Benzing2022GradientDO} imply that it can be interesting to explore the use of even simpler approximations of the FIM. More precisely, after approximating the FIM by a block diagonal matrix as in KFAC, one can further approximate each full diagonal block by an inner sub-blocks diagonal matrix (see for instance \cite{MiniBlockFisher2022}). This approach will save computational time and probably maintain the same level of optimization performance.

\bibliography{DataSciences}

\begin{thebibliography}{10}
\providecommand{\MR}{\relax\unskip\space MR }
\providecommand{\url}[1]{\normalfont{#1}}
\providecommand{\urlprefix}{Available at }

\bibitem{Amari1998}
S.I. Amari, \emph{Natural gradient works efficiently in learning}, Neur.
  Comput. 10 (1998), pp. 251--276.

\bibitem{AmariNagaoka2000}
S.I. Amari and H. Nagaoka, \emph{Methods of Information Geometry}, Translations
  of Mathematical Monographs Vol. 191, American Mathematical Society,
  Providence, Rhode Island, 2000.

\bibitem{BaGrosseMartens2017}
J. Ba, R. Grosse, and J. Martens, \emph{Distributed Second-Order Optimization
  using {K}ronecker-Factored Approximations}, in \emph{5th International
  Conference on Learning Representations, Conference Track Proceedings}, 24--26
  Apr, Toulon, France. 2017.
  \urlprefix\url{https://openreview.net/forum?id=SkkTMpjex}.

\bibitem{MiniBlockFisher2022}
A. Bahamou, D. Goldfarb, and Y. Ren, \emph{A mini-block fisher method for deep
  neural networks} (2022). \urlprefix\url{https://arxiv.org/abs/2202.04124}.

\bibitem{Bahdanau2014}
D. Bahdanau, K. Cho, and Y. Bengio, \emph{Neural machine translation by jointly
  learning to align and translate}, in \emph{3rd International Conference on
  Learning Representations}, Y. Bengio and Y. LeCun, eds., May, San Diego,
  California. 2015. \urlprefix\url{https://arxiv.org/abs/1409.0473}.

\bibitem{Benzing2022GradientDO}
F. Benzing, \emph{Gradient Descent on Neurons and its Link to Approximate
  Second-Order Optimization}, in \emph{Proceedings of the 39th International
  Conference on Machine Learning}, Vol. 162, Baltimore, Maryland. 2022.
  \urlprefix\url{https://proceedings.mlr.press/v162/benzing22a/benzing22a.pdf}.

\bibitem{BRB2017}
A. Botev, H. Ritter, and D. Barber, \emph{Practical {G}auss-{N}ewton
  Optimisation for Deep Learning}, in \emph{Proceedings of the 34th
  International Conference on Machine Learning}, D. Precup and Y.W. Teh, eds.,
  Proceedings of Machine Learning Research Vol.~70, 06--11 Aug, Sydney,
  Australia. 2017, pp. 557--565.

\bibitem{broyden1970}
C.G. Broyden, \emph{The convergence of a class of double-rank minimization
  algorithms 1. {G}eneral considerations}, IMA J. Appl. Math. 6 (1970), pp.
  76--90.

\bibitem{Chellapilla2006HighPC}
K. Chellapilla, S. Puri, and P.Y. Simard, \emph{High Performance Convolutional
  Neural Networks for Document Processing}, in \emph{10th International
  Workshop on Frontiers in Handwriting Recognition}, G. Lorette, H. Bunke, and
  L. Schomaker, eds., 23--26 Oct, La Baule, France. 2006.
  \urlprefix\url{https://hal.inria.fr/inria-00112631}.

\bibitem{Jolivet2015}
V. Dolean, P. Jolivet, and F. Nataf, \emph{An Introduction to Domain
  Decomposition Methods: Algorithms, Theory, and Parallel Implementation},
  Society for Industrial and Applied Mathematics, Philadelphia, 2015,
  \urlprefix\url{https://www.ljll.math.upmc.fr/nataf/OT144DoleanJolivetNataf_full.pdf}.

\bibitem{adagrad}
J. Duchi, E. Hazan, and Y. Singer, \emph{Adaptive subgradient methods for
  online learning and stochastic optimization}, J. Mach. Learn. Res. 12 (2011),
  pp. 2121--2159.
  \urlprefix\url{https://www.jmlr.org/papers/volume12/duchi11a/duchi11a.pdf}.

\bibitem{fletcher1970}
R. Fletcher, \emph{A new approach to variable metric algorithms}, Comput. J. 13
  (1970), pp. 317--322.

\bibitem{TKFAC}
K.X. Gao, X.L. Liu, Z.H. Huang, M. Wang, Z. Wang, D. Xu, and F. Yu, \emph{A
  trace-restricted {K}ronecker-factored approximation to natural gradient}
  (2020). \urlprefix\url{https://arxiv.org/abs/2011.10741}.

\bibitem{Gehring2017}
J. Gehring, M. Auli, D. Grangier, D. Yarats, and Y.N. Dauphin,
  \emph{Convolutional sequence to sequence learning}, in \emph{Proceedings of
  the 34th International Conference on Machine Learning}, D. Precup and Y.W.
  Teh, eds., Vol.~70, 06--11 Aug, Sydney, Australia. 2017, pp. 1243--1252.
  \urlprefix\url{https://proceedings.mlr.press/v70/gehring17a.html}.

\bibitem{GLBBV2018}
T. George, C. Laurent, X. Bouthillier, N. Ballas, and P. Vincent, \emph{Fast
  approximate natural gradient descent in a {K}ronecker factored eigenbasis},
  in \emph{Advances in Neural Information Processing Systems 31}, S. Bengio, H.
  Wallach, H. Larochelle, K. Grauman, N. Cesa-Bianchi, and R. Garnett, eds.,
  Curran Associates, Inc., Montr\'eal, Canada,  2018, pp. 9550--9560.
  \urlprefix\url{http://papers.nips.cc/paper/8164-fast-approximate-natural-gradient-descent-in-a-kronecker-factored-eigenbasis.pdf}.

\bibitem{goldfard1970}
D. Goldfarb, \emph{A family of variable-metric methods derived by variational
  means}, Math. Comp. 24 (1970), pp. 23--26.

\bibitem{GRB2020}
D. Goldfarb, Y. Ren, and A. Bahamou, \emph{Practical quasi-{N}ewton methods for
  training deep neural networks}, arXiv:2006.08877 (2020).
  \urlprefix\url{https://arxiv.org/pdf/2006.08877.pdf}.

\bibitem{GrosseMartens2016}
R. Grosse and J. Martens, \emph{A {K}ronecker-factored approximate {F}isher
  matrix for convolution layers}, in \emph{Proceedings of the 33rd
  International Conference on Machine Learning}, M.F. Balcan and K.Q.
  Weinberger, eds., Vol.~48, 19--24 Jun, New York. 2016, pp. 573--582.
  \urlprefix\url{http://proceedings.mlr.press/v48/grosse16.html}.

\bibitem{HeEtal2016}
K. He, X. Zhang, S. Ren, and J. Sun, \emph{Deep Residual Learning for Image
  Recognition}, in \emph{Proceedings of the 2016 IEEE Conference on Computer
  Vision and Pattern Recognition}, 27--30 Jun, Las Vegas, Nevada. 2016, pp.
  770--778.

\bibitem{Heskes2000}
T. Heskes, \emph{On ``natural'' learning and pruning in multilayered
  perceptrons}, Neur. Comput. 12 (2000), pp. 881--901.

\bibitem{KingmaBa2015}
D.P. Kingma and J. Ba, \emph{Adam: {A} Method for Stochastic Optimization}, in
  \emph{Proceedings of the 3rd International Conference on Learning
  Representations}, Y. Bengio and Y. LeCun, eds., 7--9 May, San Diego,
  California. 2015. \urlprefix\url{http://arxiv.org/abs/1412.6980}.

\bibitem{Koroko-et-al}
A. Koroko, A. Anciaux-Sedrakian, I. Ben~Gharbia, V. Gar{\`{e}}s, M. Haddou, and
  Q.H. Tran, \emph{Efficient approximations of the {F}isher matrix in neural
  networks using {K}ronecker product singular value decomposition},
  arXiv:2201.10285 (2022). \urlprefix\url{https://arxiv.org/abs/2201.10285}.

\bibitem{Krizhevsky2009LearningML}
A. Krizhevsky, \emph{Learning multiple layers of features from tiny images},
  Tech. {R}ep., University of Toronto, Toronto, Ontario,  2009.
  \urlprefix\url{https://www.cs.toronto.edu/~kriz/cifar.html}.

\bibitem{Krizhevsky2012}
A. Krizhevsky, I. Sutskever, and G.E. Hinton, \emph{ImageNet Classification
  with Deep Convolutional Neural Networks}, in \emph{Advances in Neural
  Information Processing System 25}, F. Pereira, C.J.C. Burges, L. Bottou, and
  K.Q. Weinberger, eds., 03--08 Dec, Lake Tahoe, California. Curran Associates,
  Inc., 2012, pp. 1097--1105.
  \urlprefix\url{https://proceedings.neurips.cc/paper/2012/file/c399862d3b9d6b76c8436e924a68c45b-Paper.pdf}.

\bibitem{LRMB2008}
N. Le~Roux, P.A. Manzagol, and Y. Bengio, \emph{Topmoumoute online natural
  gradient algorithm}, in \emph{Proceedings of the 20th International
  Conference on Neural Information Processing Systems}, J.C. Platt, D. Koller,
  Y. Singer, and S.T. Roweis, eds., 03--06 Dec, Vancouver, Canada. 2007, pp.
  849--856.

\bibitem{nocedal1989}
D.C. Liu and J. Nocedal, \emph{On the limited memory {BFGS} method for large
  scale optimization}, Math. Prog. 45 (1989), pp. 503--528.

\bibitem{martens2010}
J. Martens, \emph{Deep learning via {H}essian-free optimization}, in
  \emph{Proceedings of the 27th International Conference on Machine Learning},
  Vol.~27, 21--24 Jun, Haifa, Israel. 2010, pp. 735--742.

\bibitem{Martens2014}
J. Martens, \emph{New insights and perspectives on the natural gradient
  method}, arXiv:1412.1193 (2014).
  \urlprefix\url{https://arxiv.org/abs/1412.1193}.

\bibitem{martensthesis}
J. Martens, \emph{Second-order optimization for neural networks}, Ph.D. diss.,
  University of Toronto, Ontario, Canada,  2016.
  \urlprefix\url{http://hdl.handle.net/1807/71732}.

\bibitem{MBJ2018}
J. Martens, J. Ba, and M. Johnson, \emph{Kronecker-factored curvature
  approximations for recurrent neural networks}, in \emph{Proceedings of the
  6th International Conference on Learning Representations}, 30 Apr--3 May,
  Vancouver, Canada. 2018.
  \urlprefix\url{https://openreview.net/forum?id=HyMTkQZAb}.

\bibitem{MartensGrosse2015}
J. Martens and R. Grosse, \emph{Optimizing neural networks with
  {K}ronecker-factored approximate curvature}, in \emph{Proceedings of the 32nd
  International Conference on Machine Learning}, Vol.~37, 06--11 Jul, Lille,
  France. 2015, pp. 2408--2417.
  \urlprefix\url{http://proceedings.mlr.press/v37/martens15.html}.

\bibitem{Nataf2011}
F. Nataf, H. Xiang, V. Dolean, and N. Spillane, \emph{A coarse space
  construction based on local {D}irichlet-to-{N}eumann maps}, SIAM J. Sci.
  Comput. 33 (2011), pp. 1623--1642.

\bibitem{Nesterov1983AMF}
Y.E. Nesterov, \emph{A method for solving the convex programming problem with
  convergence rate ${O}(1/k^2)$}, Dokl. Akad. Nauk SSSR 269 (1983), pp.
  543--547. \urlprefix\url{http://mi.mathnet.ru/dan4600}.

\bibitem{Netzer2011ReadingDI}
Y. Netzer, T. Wang, A. Coates, A. Bissacco, B. Wu, and A. Ng, \emph{Reading
  Digits in Natural Images with Unsupervised Feature Learning}, in \emph{NIPS
  Workshop on Deep Learning and Unsupervised Feature Learning}, December,
  Granada. 2011.
  \urlprefix\url{http://ufldl.stanford.edu/housenumbers/nips2011_housenumbers.pdf}.

\bibitem{Nicolaides1987DeflationOC}
R.A. Nicolaides, \emph{Deflation of conjugate gradients with applications to
  boundary value problems}, SIAM J. Numer. Anal. 24 (1987), pp. 355--365.

\bibitem{olivier2015}
Y. Ollivier, \emph{Riemannian metrics for neural networks {I}: feedforward
  networks}, Inform. Infer. 4 (2015), pp. 108--153.

\bibitem{OTUNYM2019}
K. Osawa, Y. Tsuji, Y. Ueno, A. Naruse, R. Yokota, and S. Matsuoka,
  \emph{Large-Scale Distributed Second-Order Optimization Using
  {K}ronecker-Factored Approximate Curvature for Deep Convolutional Neural
  Networks}, in \emph{Proceedings of the 2019 IEEE Conference on Computer
  Vision and Pattern Recognition}, 15--20 Jun, Long Beach, California. 2019,
  pp. 12359--12367.
  \urlprefix\url{http://openaccess.thecvf.com/content_CVPR_2019/html/Osawa_Large-Scale_Distributed_Second-Order_Optimization_Using_Kronecker-Factored_Approximate_Curvature_for_Deep_CVPR_2019_paper.html}.

\bibitem{PascanuBengio2013}
R. Pascanu and Y. Bengio, \emph{Revisiting natural gradient for deep networks},
  arXiv preprint arXiv:1301.3584 (2013).
  \urlprefix\url{https://arxiv.org/abs/1301.3584v4}.

\bibitem{pytorch}
A. Paszke, S. Gross, F. Massa, A. Lerer, J. Bradbury, G. Chanan, T. Killeen, Z.
  Lin, N. Gimelshein, L. Antiga, A. Desmaison, A. Kopf, E. Yang, Z. DeVito, M.
  Raison, A. Tejani, S. Chilamkurthy, B. Steiner, L. Fang, J. Bai, and S.
  Chintala, \emph{Py{T}orch: An Imperative Style, High-Performance Deep
  Learning Library}, in \emph{Advances in Neural Information Processing Systems
  32}, H. Wallach, H. Larochelle, A. Beygelzimer, F. d'Alch\'{e}  Buc, E. Fox,
  and R. Garnett, eds., 08--14 Dec, Vancouver, Canada. Curran Associates, Inc.,
  2019, pp. 8026--8037.
  \urlprefix\url{https://proceedings.neurips.cc/paper/2019/hash/bdbca288fee7f92f2bfa9f7012727740-Abstract.html}.

\bibitem{Polyak1964SomeMO}
B. Polyak, \emph{Some methods of speeding up the convergence of iteration
  methods}, USSR Comput. Math. Math. Phys. 4 (1964), pp. 1--17.

\bibitem{PZK2014}
D. Povey, X. Zhang, and S. Khudanpur, \emph{Parallel training of {DNN}s with
  natural gradient and parameter averaging}, arXiv:1410.7455 (2014).
  \urlprefix\url{https://arxiv.org/abs/1410.7455}.

\bibitem{ritter2018a}
H. Ritter, A. Botev, and D. Barber, \emph{A Scalable {L}aplace Approximation
  for Neural Networks}, in \emph{6th International Conference on Learning
  Representations}, ICLR Conference Track Proceedings Vol.~6, Vancouver,
  Canada. 2018. \urlprefix\url{https://openreview.net/forum?id=Skdvd2xAZ}.

\bibitem{RobbinsMonro1951}
H. Robbins and S. Monro, \emph{A stochastic approximation method}, Ann. Math.
  Statist. 22 (1951), pp. 400--407.
  \urlprefix\url{https://www.jstor.org/stable/2236626}.

\bibitem{sak2014}
H. Sak, A. Senior, and F. Beaufays, \emph{Long Short-Term Memory Recurrent
  Neural Network Architectures for Large Scale Acoustic Modeling}, in
  \emph{15th Annual Conference of the International Speech Communication
  Association. Celebrating the Diversity of Spoken Languages}, H. Li and P.
  Ching, eds., 14--18 Sep, Singapore. Curran Associates, Inc., 2014, pp.
  338--342. \urlprefix\url{https://arxiv.org/abs/1402.1128}.

\bibitem{Schraudolph2002}
N.N. Schraudolph, \emph{Fast curvature matrix-vector products for second-order
  gradient descent}, Neur. Comput. 14 (2002), pp. 1723--1738.

\bibitem{Sercu2016}
T. Sercu, C. Puhrsch, B. Kingsbury, and Y. Le{C}un, \emph{Very deep
  multilingual convolutional neural networks for {LVCSR}}, in \emph{2016 IEEE
  International Conference on Acoustics, Speech and Signal Processing}, 20--25
  Mar, Shanghai, China. 2016, pp. 4955--4959.

\bibitem{shanno1970}
D.F. Shanno, \emph{Conditioning of quasi-{N}ewton methods for function
  minimization}, Math. Comp. 24 (1970), pp. 647--656.

\bibitem{sluskeretal2013}
I. Sutskever, J. Martens, G. Dahl, and G. Hinton, \emph{On the importance of
  initialization and momentum in deep learning}, in \emph{Proceedings of the
  30th International Conference on Machine Learning}, S. Dasgupta and D.
  McAllester, eds., Proceedings of Machine Learning Research Vol.~28, 17--19
  Jun, Atlanta, Georgia. 2013, pp. 1139--1147.
  \urlprefix\url{https://proceedings.mlr.press/v28/sutskever13.html}.

\bibitem{rmsprop}
T. Tieleman and G. Hinton, \emph{Lecture 6.5 {RMSP}rop: Divide the gradient by
  a running average of its recent magnitude}, COURSERA: Neural Networks for
  Machine Learning 4 (2012), pp. 26--31.

\bibitem{twolevels}
N. Tselepidis, J. Kohler, and A. Orvieto, \emph{Two-Level {K-FAC}
  Preconditioning for Deep Learning}, in \emph{12th Annual Workshop on
  Optimization for Machine Learning}, online. 2020.
  \urlprefix\url{https://arxiv.org/abs/2011.00573}.

\bibitem{kfacrl}
Y. Wu, E. Mansimov, R.B. Grosse, S. Liao, and J. Ba, \emph{Scalable
  trust-region method for deep reinforcement learning using
  {K}ronecker-factored approximation}, in \emph{Proceedings of the 31st
  International Conference on Neural Information Processing Systems}, I. Guyon,
  U.V. Luxburg, S. Bengio, H. Wallach, R. Fergus, S. Vishwanathan, and R.
  Garnett, eds., 04--09 Dec, Long Beach, California. Curran Associates, Inc.,
  2017, pp. 5285--5294.
  \urlprefix\url{https://proceedings.neurips.cc/paper/2017/file/361440528766bbaaaa1901845cf4152b-Paper.pdf}.

\bibitem{xiao2017/online}
H. Xiao, K. Rasul, and R. Vollgraf, \emph{Fashion-{MNIST}: a novel image
  dataset for benchmarking machine learning algorithms} (2017).
  \urlprefix\url{https://arxiv.org/abs/1708.07747}.

\bibitem{kfacbay}
G. Zhang, S. Sun, D. Duvenaud, and R. Grosse, \emph{Noisy Natural Gradient as
  Variational Inference}, in \emph{Proceedings of the 35th International
  Conference on Machine Learning}, J. Dy and A. Krause, eds., Proceedings of
  Machine Learning Research Vol.~80, 10--15 Jul. 2018, pp. 9308--9321.
  \urlprefix\url{https://proceedings.mlr.press/v80/zhang18l.html}.

\bibitem{ZangGrosse2018}
G. Zhang, C. Wang, B. Xu, and R. Grosse, \emph{Three mechanisms of weight decay
  regularization} (2018). \urlprefix\url{https://arxiv.org/abs/1810.12281}.

\bibitem{Zhang2019}
J. Zhang, \emph{Gradient descent based optimization algorithms for deep
  learning models training}  (2019).
  \urlprefix\url{https://arxiv.org/abs/1903.03614}.

\end{thebibliography}
\bibliographystyle{tfs}

\appendix

\section{Efficient computation of $F_{\bullet} u$ and $F_{\text{coarse}}$}\label{appendix}
\subsection{Notations}
We consider the same network architectures (MLP and CNN) and notations introduced in \S\ref{sec:preliminaries}. For two matrices $A$ et $B$ of same sizes, $A\odot B$ denotes the Hadamard (element-wise) product of $A$ and $B$. We will also write $\left<u,v\right>$ for the inner (dot) product between vectors $u$ and $v$. We recall that ``vec'' is the operator that turns a matrix into a vector by stacking its columns together and ``MAT,'' the converse of ``vec,'' turns a vector into a matrix.

For a big vector $u\in\mathbb{R}^p$, where $p$ is the number of parameters contained in $\theta$, the notation $u[i]\in \mathbb{R}^{p_i}$ stands for the part of $u$ corresponding to layer $i$, whose number of parameters is $p_i$. For example, if $u =\theta = [\text{vec}(W_1)^T,\text{vec}(W_2)^T, \hdots, \text{vec}(W_{\ell})^T]^T$, then $u[i]=\text{vec}(W_i)$ for all ${i} \in \{1,\hdots,\ell\}$.

\subsection{Computation of $F_{\bullet}u$}\label{sse:appFu}

In view of the regularization \eqref{eq:FIMreg}, it is plain that
\begin{equation}
F_{\bullet} u = Fu + \lambda u.
\end{equation}
Since the regularization term does not cause any trouble, we only have to deal with $Fu$. It is notoriously knwon that the matrix-vector product involving the Fisher matrix can be carried out
without explicitly forming $F$ thanks to an algorithm by Schraudolph \cite{Schraudolph2002}. However, this approach requires to perform additional forward/backward passes. Here, we present an efficient way to evaluate $Fu$ by re-using the quantities computed during the traditional backward/forward pass.

Suppose that we have a mini-batch $\mathcal{B}=\{(x_1,y_1),\hdots,(x_B,y_B)\}$ where targets $y_i$'s are sampled from the model predictive distribution $P_{y|x}(\theta)$. Then, $F$ is computed using a Monte Carlo estimation, i.e.,
\begin{equation}
F = [\mathcal{D}\theta(\mathcal{D}\theta)^T] \approx \frac{1}{B}\sum_{b=1}^B(\mathcal{D}\theta)^{(b)}((\mathcal{D}\theta)^{(b)})^T=\frac{1}{B}JJ^T,
\end{equation}
where $J\in \mathbb{R}^{p\times B}$ is the matrix whose $b$-th column is $(\mathcal{D}\theta)^{(b)}$. Thus,
\begin{equation}
Fu=\frac{1}{B}JJ^Tu.
\end{equation}
The computation of $Fu$ can therefore be divided into two steps: (1) matrix-vector product $v=J^Tu$; (2) matrix-vector product $\frac{1}{B}Jv$.

\paragraph*{Step 1: multiplying by $J^T$.} Since $J^T \in \mathbb{R}^{B\times p}$ and $u\in \mathbb{R}^{p}$, we have $v=J^Tu \in\mathbb{R}^B$. For all $b \in \{1,\hdots,B\}$, the $b$-th entry $v_b$ of $v$ is none other than the dot product between the $b$-th column $(\mathcal{D}\theta)^{(b)}$ of $J$ and $u$. This dot product can be split into layer-wise dot products and then summed up together. Formally, we have
\begin{equation}
    v_b = \big\langle (\mathcal{D}\theta)^{(b)},u \big\rangle = \sum_{i=1}^{\ell} \big\langle \! \vecteur((\mathcal{D}W_i)^{(b)}),u[i] \big\rangle.
\end{equation}
We now distinguish two cases according to the type of layer $i$.
\begin{enumerate}
\item If layer $i$ is an MLP, we have $\mathcal{D}W_{i}= g_{i}\Bar{a}_{i-1}^T$ and then
\begin{align}
    \big<\!\vecteur((\mathcal{D}W_i)^{(b)}),u_{[i]}\big> &= \big<\!\vecteur(g_i^{(b)}(\Bar{a}_{i-1}^{(b)})^T),u_{[i]} \big> \nonumber \\
    & = \big<\Bar{a}_{i-1}^{(b)}\otimes g_i^{(b)},u_{[i]}\big> \nonumber\\
    & = \big((\Bar{a}_{i-1}^{(b)})^T\otimes (g_i^{(b)})^T\big)u_{[i]} \nonumber\\
    & = (g_i^{(b)})^T \text{MAT}(u_{[i]}) \Bar{a}_{i-1}^{(b)}.  
\end{align}
\item If layer $i$ is a CNN layer, we have 
$\mathcal{D}W_{i}=\sum_{t=1}^{T_i} g_{i,t} \Bar{a}_{i-1,t}^T$
and therefore
\begin{align}
   \big< \! \vecteur((\mathcal{D}W_i)^{(b)}),u_{[i]} \big> & = \bigg< \!\! \vecteur \bigg( \sum_{t=1}^{T_i}g_{i,t}^{(b)}(\Bar{a}_{i-1,t}^{(b)})^T \bigg) ,u_{[i]} \bigg> \nonumber\\
   & = \bigg<\sum_{t=1}^{T_i}\vecteur(g_{i,t}^{(b)}(\Bar{a}_{i-1,t}^{(b)})^T),u_{[i]} \bigg> \nonumber\\
  & = \sum_{t=1}^{T_i}(g_{i,t}^{(b)})^T\text{MAT}(u_{[i]})\Bar{a}_{i-1,t}^{(b)}.  
\end{align}
\end{enumerate}

\paragraph*{Step 2: multiplying by $J$.} Here, we detail how to compute $Jv$, but one should not forget to multiply the result by the scaling factor $1/B$. Let $v \in \mathbb{R}^B$. $Jv$ is a weighted sum of the columns of $J$ with the weight coefficients corresponding to the entries of $v$. In other words,
\begin{equation}
Jv = \sum_{b=1}^Bv_b(\mathcal{D}\theta)^{(b)}.
\end{equation}
For all $i \in \{1,\hdots,\ell\}$, the part of $Jv$ corresponding to layer $i$ is a linear combination of those of columns of $J$ corresponding to layer $i$, that is,
\begin{equation}
(Jv)[i] = \sum_{b=1}^Bv_b\vecteur((\mathcal{D}W_i)^{(b)}) .
\end{equation}
As in step 1, we have to distinguish two cases according the type of layer $i$.
\begin{enumerate}
\item If layer $i$ is an MLP, we have 
\begin{equation}
   (Jv)[i] = \sum_{b=1}^Bv_b\vecteur(g_i^{(b)}(\Bar{a}_{i-1}^{(b)})^T) = \vecteur \Big(\sum_{b=1}^Bv_bg_i^{(b)}(\Bar{a}_{i-1}^{(b)})^T \Big),
\end{equation}
and then 
\begin{equation}\label{eq:MATJvi}
    \MAT \big((Jv)_{[i]} \big) = \sum_{b=1}^Bv_bg_i^{(b)}(\Bar{a}_{i-1}^{(b)})^T
    = \big[ (\mathds{1}v^T)\odot \hat{\mathcal{G}}_i \big]\hat{\mathcal{A}}_{i-1}^T,
\end{equation}
where $\mathds{1} \in \mathbb{R}^{d_i}$ is a vector of all one's,
$\hat{\mathcal{A}}_{i-1}=(\Bar{a}_{i-1}^{(1)},\hdots,\Bar{a}_{i-1}^{(B)}) \in \mathbb{R}^{d_{i-1}\times B}$ is the matrix of activations, and
$\hat{\mathcal{G}}_i=(g_i^{(1)},\hdots,g_i^{(B)}) \in \mathbb{R}^{d_i\times B}$ is the matrix containing pre-activations derivatives.
Note that the last equality in \eqref{eq:MATJvi} is due to the fact that for two matrices $A=[A_1,\hdots,A_m] $ and $B=[B_1,\hdots,B_m]$, we have
\begin{equation}
    AB^T=\sum_kA_kB_k^T.
\end{equation}
\item If layer $i$ is a CNN, we have
\begin{align}
(Jv)_{[i]} &= \sum_{b=1}^Bv_b \vecteur \Big( \sum_{t=1}^{T_i}g_{i,t}^{(b)}(\Bar{a}_{i-1,t}^{(b)})^T \Big) \nonumber\\
& =  \sum_{b=1}^B \vecteur \Big( \sum_{t=1}^{T_i}v_bg_{i,t}^{(b)}(\Bar{a}_{i-1,t}^{(b)})^T \Big),
\end{align}
and then 
\begin{align}
    \MAT\big((Jv)_{[i]} \big) &= \sum_{b=1}^B v_b\sum_{t=1}^{T_i}g_{i,t}^{(b)}(\Bar{a}_{i-1,t}^{(b)})^T \nonumber\\
        &= \big[ (\mathds{1}V^T)\odot \hat{\mathcal{G}}_i \big] [\![\hat{\mathcal{A}}_{i-1}]\!]^T,
\end{align}
where here $\mathds{1}$ is a vector of all one's of size $c_i$ (the number of output channels), whereas
\begin{subequations}
\begin{align}
V &= (v_1,\hdots v_1|,\hdots,|v_B,\hdots v_B)^T\\
[\![\hat{\mathcal{A}}_{i-1}]\!]  &= \big( \Bar{a}_{i-1,1}^{(1)},\hdots,\Bar{a}_{i-1,T_i}^{(1)}|\Bar{a}_{i-1,1}^{(2)},\hdots,\Bar{a}_{i-1,T_i}^{(2)}|\hdots,|\Bar{a}_{i-1,1}^{(B)},\hdots,\Bar{a}_{i-1,T_i}^{(B)} \big) ,\\
\hat{\mathcal{G}}_i &= \big( g_{i,1}^{(1)},\hdots,g_{i,T_i}^{(1)}|g_{i,1}^{(2)},\hdots,g_{i,T_i}^{(2)}|\hdots,|g_{i,1}^{(B)},\hdots,g_{i,T_i}^{(B)} \big),
\end{align}
\end{subequations}
are respectively in $\mathbb{R}^{T_iB}$ (a duplicated version of $v$), $\mathbb{R}^{({c_{i-1}\Delta_i+1})\times T_iB}$ and $\mathbb{R}^{c_i\times T_iB}$.
\end{enumerate}

\subsection{Computation of $F_{\text{coarse}}$}\label{sse:appFcoarse}
From definition \eqref{eq:Fcoarse} of the coarse operator and the a priori form \eqref{eq:formeR0T} of the coarse space, it follows that
\begin{equation}\label{eq:FcoarseIJ}
[F_{\text{coarse}}]_{i,j} = V_i^T [F_{\bullet}]_{i,j} V_j
\end{equation}
for all $(i,j) \in \{1,\hdots,\ell\}\times \{1,\hdots,\ell\}$. 
In view of the regularization \eqref{eq:FIMreg}, the entry \eqref{eq:FcoarseIJ} becomes
\begin{equation}
[F_{\text{coarse}}]_{i,j} = \begin{cases}
    \, V_i^T F_{i,j} V_j   & \; \text{if } \, i\neq j,\\
    \, V_i^T F_{i,i} V_i + \lambda V_i^T V_i & \; \text{if } \, i = j.
\end{cases}
\end{equation}
Since the regularization term $\lambda V_i^T V_i$ does not cause any trouble, we only have to deal with the elementary products $v^T F_{i,j} w$, where $v\in \mathbb{R}^{p_i}$ (a column of $V_i$) and $w\in \mathbb{R}^{p_j}$ (a column of $V_j$).
We reall that $F_{i,j} \in\mathbb{R}^{p_i\times p_j}$ is computed using a Monte Carlo estimation on a mini-batch, i.e.,
\begin{equation}
F_{i,j} \approx\frac{1}{B}\sum_{b=1}^B\vecteur(\mathcal{D}W_i)^{(b)}(\vecteur(\mathcal{D}W_j)^{(b)})^T =\frac{1}{B} J_iJ_j^T,
\end{equation}
with
\begin{subequations}
\begin{alignat}{2}
J_i & = \big(\! \vecteur(\mathcal{D}W_i)^{(1)},\vecteur(\mathcal{D}W_i)^{(2)},\hdots,\vecteur(\mathcal{D}W_i)^{(B)}\big) & & \in \mathbb{R}^{p_i\times B} ,\\
J_j & = \big(\! \vecteur(\mathcal{D}W_j)^{(1)},\vecteur(\mathcal{D}W_j)^{(2)},\hdots,\vecteur(\mathcal{D}W_j)^{(B)} \big) & & \in \mathbb{R}^{p_j\times B}.
\end{alignat}
\end{subequations}
Then, $v^TF_{i,j}w$ is given by
\begin{equation}
v^T F_{i,j} w =\frac{1}{B} v^T J_i J_j^T w.
\end{equation}

The computation of $v^T F_{i,j} w$ can therefore be performed in three steps: (1) matrix-vector product $\mu=J_j^T w$;  (2) matrix-vector product $\gamma=J_i\mu$; (3) dot product $\frac{1}{B} \langle v,\gamma \rangle$. 
 The computation of $\mu=J_j^Tv$ is done in the same way as $J^Tu$ in the previous subsection \S\ref{sse:appFu}. The only difference is that here we do not sum over layers. Likewise, the computation of $\gamma=J_i\mu$ is done exactly in the same way as $(Jv)[i]$ in \S\ref{sse:appFu}. Finally, step 3 is the classical dot product and is straightforward. 
\end{document}